\documentclass{article}
\usepackage[utf8]{inputenc}
\usepackage{enumitem}
\usepackage[english]{babel}
\usepackage{amsthm}
\usepackage{caption}
\usepackage{graphicx}
\usepackage{amssymb}
\usepackage{mathtools}
\usepackage{setspace}
\usepackage{xcolor}
\usepackage{subcaption}
\usepackage{caption}
\usepackage[ruled,vlined]{algorithm2e}
\usepackage{jmlr2e}
\usepackage{changepage}
\usepackage{wrapfig}
\usepackage{siunitx}
\usepackage{xr}

\def\cX{{\mathcal X}}
\def\k{{\kappa}}
\def\G{{\Gamma}}
\def\t{{\tau}}
\def\ul{\underline}

\newtheorem{dfn}{Definition}
\newtheorem{ntn}[dfn]{Notation}
\newtheorem{rmk}[dfn]{Remark}

\ShortHeadings{Topological Convolutional Neural Networks}{authors}
\firstpageno{1}

\begin{document}

\title{Topological Deep Learning}

\author{%
  \name Ephy R. Love \email elove4@vols.utk.edu \\
  \addr Bredesen Center DSE\\
  University of Tennessee\\
   Knoxville, TN 37996, USA
  \AND
  \name Benjamin Filippenko \email benfilip@stanford.edu\\
  \addr Department of Mathematics\\
  Stanford University\\
   Stanford, CA 94305, USA
  \AND
  \name Vasileios Maroulas \email vmaroula@utk.edu\\
  \addr Department of Mathematics\\
  University of Tennessee\\
   Knoxville, TN 37996, USA
  \AND
  \name Gunnar Carlsson \email Gunnar@math.stanford.edu\\
  \addr Department of Mathematics\\
  Stanford University\\
   Stanford, CA 94305, USA
}

\editor{Genevera Allen, Sayan Mukherjee, Boaz Nadler}

\date{February 2020}

\maketitle

\begin{abstract}%
This work introduces the Topological CNN (TCNN), which encompasses several topologically defined convolutional methods. Manifolds with important relationships to the natural image space are used to parameterize image filters which are used as convolutional weights in a TCNN. These manifolds also parameterize slices in layers of a TCNN across which the weights are localized. We show evidence that TCNNs learn faster, on less data, with fewer learned parameters, and with greater generalizability and interpretability than conventional CNNs. We introduce and explore TCNN layers for both image and video data. We propose extensions to 3D images and 3D video.
\end{abstract}
\begin{keywords}
machine learning, convolutional neural network, topology, topological data analysis, image and video classification
\end{keywords}

\tableofcontents

\section{Introduction}

It was observed in \cite{mnist98} that one motivation for the construction of Convolutional Neural Networks (CNNs) was that they permit a sparsification based on the geometry of the space of features.  In  the case of convolutional neural networks for images, the geometry used was that of a two-dimensional grid of features, in this case pixels. M. Robinson has also pointed out the importance of geometries or topologies on spaces of features, coining the term {\em topological signal processing} to describe this notion \cite{robinsontps}.   In this paper, we study  a space of image filters closely related to a subfamily of the Gabor filters, whose geometry is that of a well known geometric object, the {\em Klein bottle}\footnote{See Figure~\ref{fig:KOLconnections} for an image of a Klein Bottle immersed in $\mathbb{R}^3$.}.  These filters augment the feature set in the data, but they can also be used to construct analogues of convolutional neural networks with improved performance on a number of measures of performance.  The method uses a  discretization (in the form of a graph)  of the Klein bottle as a template for new layers, which produce additional sparsification.

We implement the use of the Klein bottle geometry and its image filters via additional structure on convolutional layers. We call neural networks with these layers {\bf Topological Convolutional Neural Networks (TCNNs)}. We perform experiments on image and video data. The results show significant improvement in TCNNs compared to conventional CNNs with respect to various metrics.

Deep neural network (NN) architectures are the preeminent tools for many image classification tasks, since they are capable of distinguishing between a large number of classes with a high degree of precision and accuracy \cite{guo2016deep}. CNNs are components of the most commonly used neural network architecture for image classification, e.g. see \cite{he2016deep,rawat2017deep,nips20124824}. The characterizing property of CNNs is their use of convolutional layers which take advantage of the 2-dimensional topology of an image to sparsify a fully connected network and employ weight sharing across slices. Each convolutional layer in a CNN assembles spatially local features, e.g.\ textures, lines, and edges, into complex global features, e.g.\ the location and classification of objects. CNNs are also used to classify videos; see e.g.\ \cite{UCF101}, \cite{1334462}, and \cite{ActionsAsSpaceTimeShapes_pami07}. CNNs have several major drawbacks including that the models are difficult to interpret, require large datasets, and often do not generalize well to new data~\cite{zheng2018improvement}. It has been demonstrated that as CNNs grow in size and complexity they often do not enjoy a proportional increase in utility \cite{he2016deep}. This suggests that bigger, deeper models alone will not continue to advance image classification.

The topological structure in the layers of a TCNN is inspired from prior research on natural image data with topological data analysis (TDA). Small patches in natural images cluster around a Klein bottle embedded in the space of patches \cite{carlsson_local_2008}. The patches corresponding to this Klein bottle are essentially edges, lines, and interpolations between those; see top panels of Figure~\ref{fig:featureActivationsOn5}(b)(c). It has been shown through TDA that CNN weights often arrive at these same Klein bottle patches after training \cite{carlsson_topological_2018}. The key idea of TCNNs is to use the empirically discovered Klein bottle, and the image patches it represents, directly in the structure of the convolutional layers. For example, the Klein bottle image patches are used in TCNNs as convolutional filters that are fixed during training. There is an analogue of the Klein bottle for video data. Because of symmetries present in the geometric models of feature spaces, there are generalized notions of weight sharing which encode invariances under rotation and black-white reversal.

The method is not simply an addition of features to particular data sets, but is in fact a general methodology that can be applied to all image or video data sets.  In the case of images, it builds in the notion that edges and lines are important features for any image data set, and that this notion should be included and accounted for in the architecture for any kind of image data analysis. TDA contains a powerful set of methods that can be used to discover these latent manifolds on which data sit \cite{bubenik_statistical_2015,chazal2017robust,JMLR_maroulas,sgouralis_bayesian_2017}. These methods can then be used to inform the parametrization of the TCNN.

TCNNs are composed of two new types of convolutional layers which construct topological features and restrict convolutions based on embeddings of topological manifolds into the space of images (and similarly for video). TCNNs are inherently easier to interpret than CNNs, since in one type of layer (Circle Features layer and Klein Features layer; Definition~\ref{def:fixedweightlayers2d}) the local convolutional kernels are easily interpreted as points on the Klein bottle, all of which have clear visual meaning (e.g.\ edges and lines). Following a Klein Features layer it makes sense to insert our other type of topological layer (Circle One Layer and Klein One Layer; Definitions~\ref{def:circCorr}~ and~\ref{def:kleinCorr}), in which the 2D slices in the input and output are parameterized by a discretization of the Klein bottle and all weights between slices that are farther away than a fixed threshold distance on the Klein bottle are held at zero throughout training. This has the effect of localizing connections in this convolutional layer to slices that are nearby to each other as points on the Klein bottle. The idea of this `pruned' convolutional layer is that it aggregates the output of Klein bottle filters from the first layer that are all nearby each other. \emph{Both of these new types of topological convolutional layers can be viewed as a strict form of regularization, which explains the ability of the TCNN to more effectively generalize to new data.} We also provide versions of these layers for video and 3D convolutions.

The main points of the paper are as follows.  
\begin{enumerate}
\item{ The method provides improved performance on measures of accuracy,  speed of learning and data requirements, and generalization over standard CNNs.  }
\item{Our method can be substituted for standard convolutional neural network layers that occur in other approaches, and one should expect improvements in these other situations as well. In this paper, we compare TCNNs to traditional CNNs with the same architecture except for the modified convolutional layers. The comparatively better performance of the TCNNs provides evidence that TCNNs improve the general methodology of CNNs, with the expectation that these advantages can be combined with other methods known to be useful. For example, we use TCNN layers in a ResNet for video classification; see Section~\ref{sec:videoexperiments}. As another example, we expect state-of-the-art transfer learning results to improve when some convolutional layers in the network are replaced by TCNN layers. }
\item{Our approach suggests a general methodology for building analogues of neural networks in contexts other than static images.  We carry this out for the study of  video data.  In this case, the Klein bottle $\mathcal{K}$  is replaced by a different manifold, the so-called {\em tangent bundle} to $\mathcal{K}$, denoted by $T(\mathcal{K})$.   There are also straightforward extensions to 3D imaging and video.  The improvement afforded by these methods is expected to increase as the data complexity increases, and we find this to be the case in the passage from static images to video.    }
\item{The simple geometries of the feature spaces enable reasoning about the features to use.  In the video situation, we found that using the entire manifold was computationally infeasible, and selected certain natural submanifolds within $T(\mathcal{K})$, which allowed for the improved performance. Even in the case of static images, there are natural submanifolds of $\mathcal{K}$ which might be sufficient for the study of specific classes of images, such as line drawings, and that restricting to them would produce more efficient computation. } 
\item{There are more general methods of obtaining geometries on feature spaces, which do not require that the feature space occur among the family of manifolds already studied by mathematicians.  One such method is the so-called Mapper construction \cite{SPBG:SPBG07:091-100}.  A way to use  such structures to produce a sparsified neural network structure has been developed \cite[Section 5.3]{carlsson_topological_2018}.} The manifold methods described in this paper are encapsulated in the general framework described in Remark~\ref{rmk:generaldef} which can be applied to other domains.
\item{Because of the intuitive geometric nature of the approach, it permits additional transparency into the performance of the neural network.  }
\item{There are no pretrained components in any of the models considered in this paper. This is significant in particular because the vast majority of state-of-the-art models for classification of the UCF-101 video dataset do use pretrained components \cite{kalfaoglu2020late}, \cite{qiu2019learning}, \cite{carreira2017quo} }.
\end{enumerate}

The structure of the paper is as follows. In Section \ref{sec:background} we recall the basic structure of convolutional neural networks and we set up notation. TCNNs are introduced in Section \ref{sec:tcnn}; the version for image data is in Section \ref{subsec:2dimages}, the video version is in Section \ref{subsec:2dvideo}, and a connection with Gabor filters is explained in Section \ref{subsec:gabor}. Then in Section \ref{sec:imageexperiments} we describe experiments and results comparing TCNNs to standard CNNs on image data. In Section \ref{sec:videoexperiments} we do similar experiments on video data.

\section{Background: CNNs}\label{sec:background}
In this section we describe in detail the components of the CNN critical to the construction of the TCNN. One of the central motivations of the structure of a CNN is the desire to balance the learning of spatially local and global features. A convolutional layer can be thought of as a sparsified version of the fully connected, feed-forward (FF) layer, with additional homogeneity enforced across subsets of weights. The traditional CNN makes use of the latent image space by creating a small perceptive field with respect to the $L^\infty-$distance in which weights can be nonzero, thus sparsifying the parameter space by enforcing locality in the image. Additionally, homogeneity of weights across different local patches is enforced, further reducing the parameter space by using the global structure of the grid.

\begin{figure}[hbt!]
    \begin{center}
        \includegraphics[width=1\textwidth]{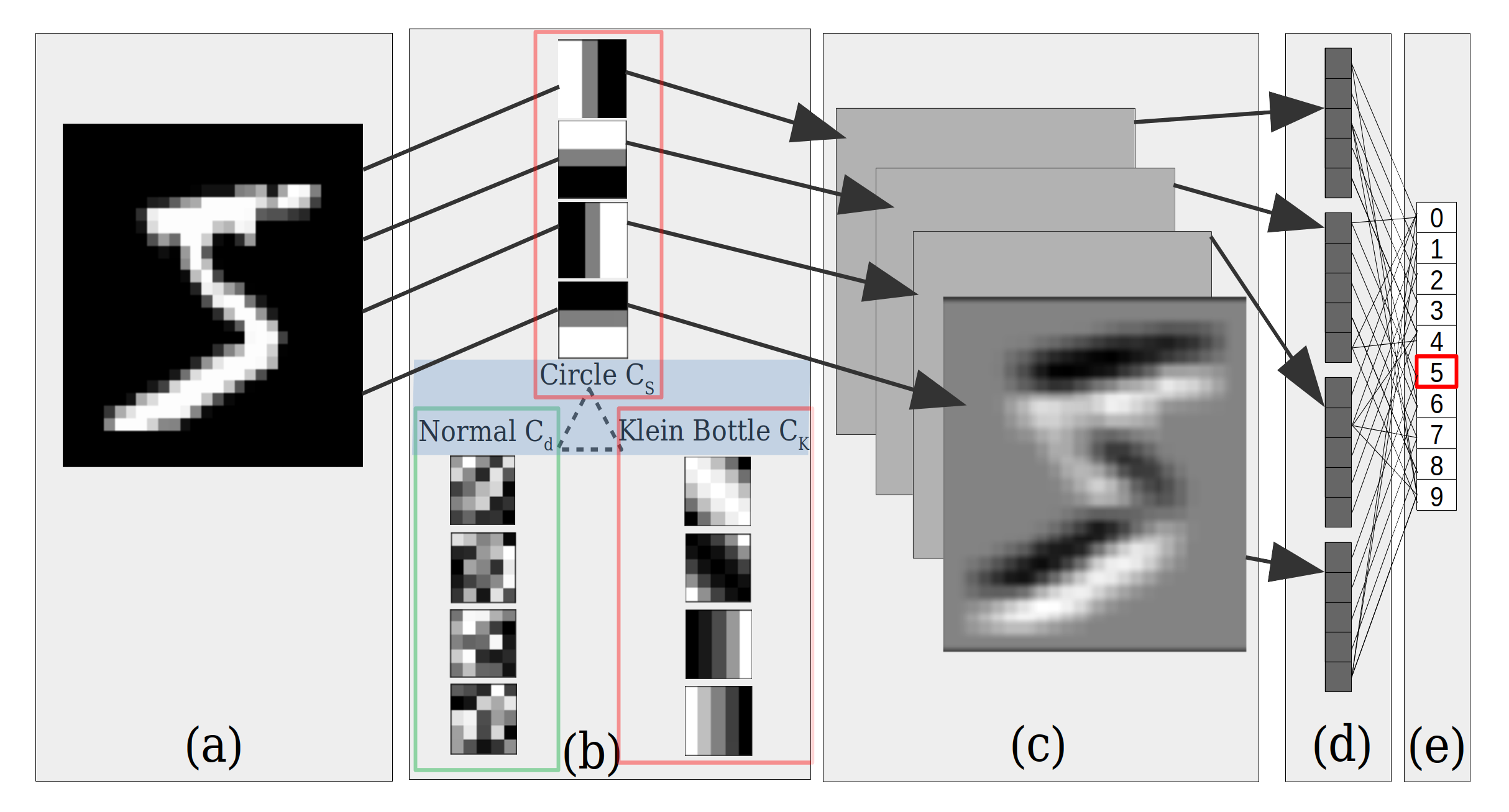}
    \end{center}
    \caption{Visual guide to CNN (green rectangle) and TCNN (red rectangles) architectures. A typical CNN layer for image classification takes in a 2-dimensional array (a) as an input image and convolves this image through a system of spatially localized filters (b) to produce multiple slices of processed data (c) called feature maps. The feature maps in (c) are flattened to a column vector in (d) and passed through a fully-connected layer to the output nodes (e) which are of cardinality equal to the number of classes in the prediction space. The TCNN modifies the typical CNN framework by specifying the weights in (b) to lie on a topological manifold such as the circle or Klein bottle. We indicate this choice by the dashed triangle in (b) and red rectangles indicating the selection of circle or Klein weights. This shows the behavior of CF and KF type TCNN layers. The COL and KOL type TCNN layers also modify the weights of a traditional CNN layer but in a different way: They localize weights with respect to the circle and Klein bottle topologies yet, as usual, instantiate them randomly and allow them to vary during training.}\label{fig:guide}
\end{figure}

To describe the structure of CNNs and TCNNs, we adopt the language in \cite{carlsson_topological_2018}, which we summarize as needed. To aid the reader, we provide an accompanying visual guide in Figure~\ref{fig:guide}.

We describe a feed forward neural network (FFNN) as a directed, acyclic graph (Definition~\ref{def:ffnndef}).

\begin{dfn} \label{def:ffnndef} A {\bf Feed Forward Neural Network (FFNN)} is a directed acyclic graph $\G$ with a vertex set $V(\G)$ satisfying the following properties:

\begin{enumerate}

\item  $V(\G)$ is decomposed as the disjoint union of its {\bf layers}
$$V(\G)=V_0(\G)\sqcup V_1(\G)...\sqcup V_r(\G).$$

\item  If $v \in V_i(\G)$, then every edge $(v,w)$ of $\G$ satisfies $w \in V_{i+1}(\G)$.

\item  For every non-initial node $w \in V_{i}(\G): i>0$, there is at least one $v \in V_{i-1}(\G)$ such that $(v,w)$ is an edge of $\G$.
 \end{enumerate}
 The vertices in $V(\G)$ are also called {\bf nodes}. For all $i$, $V_i(\G)$ consists of the {\bf nodes in layer-$\boldsymbol{i}$}. The $0^{th}$ layer $V_0(\G)$ consists of the {\bf inputs} to the neural network (Figure~\ref{fig:guide} (a)). The last layer $V_r(\G)$ consists of the {\bf outputs}. 
\end{dfn}

\begin{ntn}
Let $\G$ be a FFNN with vertex set $V(\G)$.
\begin{enumerate}
\item Often we suppress $\G$ in the notation, writing $V = V(\Gamma)$ for the set of all nodes and $V_i = V_i(\Gamma)$ for the set of nodes in layer $i$.

\item For any $v \in V$, the set $\G(v)$ consists of all $w \in V$ such that $(v,w)$ is an edge in $\G$, and the set $\G^{-1}(v)$ consists of all $w$ such that $(w,v)$ is an edge in $\G$.
\end{enumerate}
\end{ntn}

To describe the edges between nodes in successive layers, we use the notion of a {\bf correspondence} between $V_i$ and $V_{i+1}$, which is simply a subset of the product $C \subset V_i \times V_{i+1}.$  
For $v_0 \in V_i$ and $w_0 \in V_{i+1}$, we define the subsets
\begin{align*}
C(v_0) &:= \{ w \in V_{i+1} \,\, | \,\, (v_0,w) \in C \} \subset V_{i+1},\\
C^{-1}(w_0) &:= \{ v \in V_i \,\, | \,\, (v,w_0) \in C \} \subset V_i.\\
\end{align*}
Note that $C$ is determined by the subsets $C(v_0) \subset V_{i+1}$ for $v \in V_i$, and it is also determined by $C(w_0)$ for $w_0 \in V_{i+1}$. In this way, a correspondence is a generalization of a map from $V_i$ to $V_{i+1}$; Given an element in $V_i$ the correspondence provides a subset of $V_{i+1}$, and we denote correspondences by $V_i \xrightarrow{C} V_{i+1}$.

We adopt the convention that given nodes $v \in V_i, w \in V_{i+1}$, the edge $(v,w)$ is in $\G$ if and only if $(v,w) \in C.$ Note that this implies $C(v) = \Gamma(v)$. We call $C$ the {\bf edge-defining correspondence of the layer}. In this way, a FFNN specifies an edge-defining correspondence between each pair of successive layers, and conversely, choices of edge-defining correspondences between successive layers specify the edges in a FFNN. The simplest type of layer in a neural network is as follows.

\begin{dfn} \label{def:fullyconnectedlayer}
Let $V_{i+1}$ be a layer in a FFNN. We call $V_{i+1}$ a {\bf fully connected layer} if the edge-defining correspondence $C \subset V_i \times V_{i+1}$ is the entire set. In that case, we denote this correspondence \ $C_c = V_i \times V_{i+1}.$
\end{dfn}

We proceed to describe convolutional layers. We model digital images as grids indexed by $\mathbb{Z}^2$. Modifications of our constructions to finite size images will be clear. The values of the grid specifying a grayscale image are then triples $(x,y,i): x,y\in \mathbb{Z}$ where $i \in [0,1]$ is the intensity at $(x,y)$. Equivalently, an image is a map $\mathbb{Z}^2 \rightarrow [0,1]$. Similarly, videos are modeled as grids indexed by $\mathbb{Z}^3 = \mathbb{Z}^2 \times \mathbb{Z}$, where the $\mathbb{Z}^2$ are the spacial dimensions and the third dimension is time. This generalizes to grids indexed by $\mathbb{Z}^N$ for any positive integer $N$.

In a CNN, the nodes in each convolutional layer form multiple grids of the same size. We model the nodes in a convolutional layer as the product of a finite index set $\cX$ with a grid, $V_i = \cX \times \mathbb{Z}^N.$ With this notation, the graph structure of a CNN is specified as in the following definition. A convolutional layer is a sparsification of a fully connected layer which enforces locality in the grids $\mathbb{Z}^N$. For further detail, see~\cite{carlsson_topological_2018}. We explain the homogeneity restrictions on the weights in \eqref{eq:weighthomogeneity}.

\begin{dfn} \label{def:cnndef} Let $V_{i+1}$ be a layer in a FFNN. We call $V_{i+1}$ a {\bf convolutional layer} or a {\bf normal one layer (NOL)} if $V_i = \cX \times \mathbb{Z}^N$ and $V_{i+1} = \cX' \times \mathbb{Z}^N$ for some finite sets $\cX$ and $\cX'$ and a positive integer $N$, and if for some fixed {\bf threshold} $s \geq 0$ the edge-defining correspondence $C \subset V_i \times V_{i+1}$ is of the form
$$C = C_c \times C_{d,N}(s),$$
where $C_c = \cX \times \cX'$ is the fully connected correspondence and $C_{d,N}(s) \subset \mathbb{Z}^N \times \mathbb{Z}^N$ is the correspondence given by
$$C_{d,N}(s)^{-1}(\ul{x}') := \{ \ul{x} \in \mathbb{Z}^N \,\, | \,\, d_{\mathbb{Z}^N}(\ul{x},\ul{x}') \leq s \}$$
for all $\ul{x}' \in \mathbb{Z}^N$. Here, $d_{\mathbb{Z}^N}$ is the $L^{\infty}$-metric on $\mathbb{Z}^N$ defined by
$$d_{\mathbb{Z}^N}(\ul{x},\ul{x}') = \max \{|x_1 - x'_1|, \ldots, |x_N - x'_N| \}.$$

A {\bf Convolutional Neural Network (CNN)} is a FFNN such that the first layers $V_0,\ldots,V_i$ are convolutional layers and the final\footnote{Usually, there are $1$ to $3$ fully connected layers.} layers $V_{i+1},\ldots,V_{r-1}$ are fully connected.
\end{dfn}

\begin{rmk}
It is common to have pooling layers following convolutional layers in a CNN. Pooling layers downsample the size of the grids. They can be used in TCNNs in the same way and for the same purposes as traditionally used in CNNs.  We do not discuss pooling in this paper for simplicity.
\end{rmk}

The correspondence $C_{d,N}(s)^{-1}$ maps a vertex to a vertex set that is localized with respect to the threshold $s$. In a 2-dimensional image, the above definition gives the typical square convolutional filter. This constructions results in spatially meaningful graph edges.

The graph structures given in Definitions~\ref{def:ffnndef}, \ref{def:fullyconnectedlayer}, and \ref{def:cnndef} yield the skeleton of a CNN. To pass data through the network, we need a system of functions on the nodes and edges of $\G$ that pass data from layer $i-1$ to layer $i$. These functions are called {\bf activations} and {\bf weights}. The weights are real numbers $\lambda_{v,w}$ associated to each edge,
$$\Lambda=\{\lambda_{v,w} \,\, | \,\, v \in V_{i-1}, w \in V_{i}, \text{ and } (v,w) \in \G \}.$$
Let $V_{i-1} = \cX \times \mathbb{Z}^N$ and $V_i = \cX' \times \mathbb{Z}^N$, as in a CNN. Denote $v = (\kappa,\ul{x}) \in V_{i-1}$ and $w = (\kappa',\ul{x}') \in V_i$. Then the homogeneity of the weights, a characteristic of a CNN, is the translational invariance
\begin{equation} \label{eq:weighthomogeneity}
\lambda_{(\kappa,\ul{x}), (\kappa',\ul{x}')} = \lambda_{(\kappa,\ul{x} + \ul{z}), (\kappa',\ul{x}' + \ul{z})}.
\end{equation}
The activations $\mathcal{A}= \{ (u_v,f_v) \,\, | \,\, v \in V(\G) \}$ associate to each vertex $v$ a real number $u_v$ and a function $f_v : \mathbb{R}\to\mathbb{R}$. To pass data from layer $i-1$ to layer $i$ means to determine $u_w$ for $w \in V_i$ from the values $u_v, v \in V_{i-1}$ via the formula
$$u_w := f_w\bigg(\sum\limits_{v \in \G^{-1}(w)}\lambda_{v,w} \cdot u_v\bigg).$$

The {\bf activation functions} $f_v$ in neural networks usually map $x\in\mathbb{R}$ to $0<x<1$ or $0\leq x$. The output activations are a probability distribution on the output nodes of $\G$, so they are non-negative real numbers. We use the activation function ReLU, defined as $f(x)=\max(0,x)$, for all non-terminal layers, and the softmax $\sigma(x_i)=e^{x_i} (\sum\limits_j e^{x_j})^{-1}:~~ i,j\in\{1...n\}, ~~ x_i\in \mathbb{R}^+$ as the terminal layer activation function. We choose a common optimization method, adaptive moment estimation (Adam), to determine the back-propagation of our changes based on the computed gradients.

Figure~\ref{fig:guide} (a,b,c) displays an example of a weight and activation system for a convolutional layer from $V_0 = \cX(1)\times \mathbb{Z}^2$ to $V_1 = \cX(4)\times \mathbb{Z}^2$, where $\cX(\eta)$ denotes a finite set of cardinality $\eta$. The correspondence is $C_c \times C_{d,2}(2) \subset V_0 \times V_1$, where $C_c = \cX(1) \times \cX(4)$ is fully connected and $C_{d,2}(2) \subset \mathbb{Z}^2 \times \mathbb{Z}^2$ localizes connections at $L^{\infty}-$distance $2$. Panel (a) shows the input $V_0$, each weight-matrix in (b) is a vector of coefficients from $\{\lambda_{v,w}\}$, (c) shows the resulting activations in $V_1$.

\section{Topological Convolutional Layers}\label{sec:tcnn}

In Section \ref{subsec:2dimages} we introduce new types of neural network layers that form our TCNNs for 2D image classification. In Section \ref{subsec:2dvideo} we introduce layers for TCNNs used for video classification. Section \ref{subsec:gabor} demonstrates the connection between our constructions and Gabor filters.

\subsection{2D Images} \label{subsec:2dimages}
Locality in a typical convolutional neural network is a function of the $L^\infty-$distance between cells, which is specified by the correspondence $C_{d,2}$ (see Definition~\ref{def:cnndef}) in the case of a $2-$dimensional image. We add novel, topological criteria to this notion of locality through metrics on topological manifolds. The general technique is described in the following remark. 

\begin{rmk} \label{rmk:generaldef}
The layers defined in Definitions~\ref{def:kleinCorr},~\ref{def:circCorr},~\ref{dfn:6dmkc}~\ref{dfn:MVideofeatureslayer}, are all given by an edge-defining correspondence $C$ of the following general form. 

Let $M$ be a manifold and let $\cX, \cX' \subset M$ be two discretizations of $M$, meaning finite sets of points. Let $V_i = \cX \times \mathbb{Z}^N$ and $V_{i+1} = \cX' \times \mathbb{Z}^N$ be successive layers in a FFNN. Fix a threshold $s \geq 0$. Let $d$ be a metric on $M$. Define a correspondence $C(s) \subset \cX \times \cX'$ by
$$C(s)^{-1}(\k') = \{ \k \in \cX \,\, | \,\, d(\k, \k')  \leq s \}$$
for all $\k' \in \cX'$. Together with another threshold $s' \geq 0$, this defines a correspondence $C \subset V_i \times V_{i+1}$ by
$$C = C(s) \times C_{d,N}(s'),$$
where $C_{d,N}(s')$ is the convolutional correspondence from Definition~\ref{def:cnndef}. This means that
\begin{align*}
C^{-1}(\k',\ul{x}') &= C_S(s)^{-1}(\k') \times C_{d,2}(s')^{-1}(\ul{x}')\\
&= \{ (\k,x) \in \cX \times \mathbb{Z}^N \,\, | \,\, d_S(\k, \k') \leq s \text{ and } d_{\mathbb{Z}^N}(\ul{x},\ul{x}') \leq s' \}
\end{align*}
for all $(\k',\ul{x}') \in \cX \times \mathbb{Z}^N$.
\end{rmk}

The first example we give is a layer that localizes with respect to position on a circle in addition to the usual $L^\infty-$locality in a convolutional layer. Let $S^1 = \{ \k \in \mathbb{R}^2 \,\, | \,\, |\k| = 1\}$ be the unit circle in the plane $\mathbb{R}^2$. A typical discretization of $S^1$ is the set of $n$-th roots of unity $\cX = \{ e^{2\pi i k / n} \,\, | \,\, 0 \leq k \leq n-1 \}$ for some $n \geq 1$.

\begin{dfn}\label{def:circCorr}
Let $\cX, \cX' \subset S^1$ be two discretizations of the circle. Let $V_i = \cX \times \mathbb{Z}^2$ and $V_{i+1} = \cX' \times \mathbb{Z}^2$ be successive layers in a FFNN. Fix a threshold $s \geq 0$.

The {\bf circle correspondence} $C_S(s) \subset \cX \times \cX'$ is defined by
$$C_S(s)^{-1}(\k') = \{ \k \in \cX \,\, | \,\, d_S(\k, \k')  \leq s \}$$
for all $\k' \in \cX'$, where the metric $d_S$ is given by
$$d_S(\k, \k') = \cos^{-1}(\k \cdot \k') \text{ for } \k, \k' \in S^1.$$

We call $V_{i+1}$ a {\bf circle one layer (COL)} if, for some other threshold $s' \geq 0$, the edge defining correspondence $C \subset V_i \times V_{i+1}$ is of the form
$$C = C_S(s) \times C_{d,2}(s'),$$
where $C_{d,2}(s')$ is the convolutional correspondence from Definition~\ref{def:cnndef}. This means that
\begin{align*}
C^{-1}(\k',x',y') &= C_S(s)^{-1}(\k') \times C_{d,2}(s')^{-1}(x',y')\\
&= \{ (\k,x,y) \in \cX \times \mathbb{Z}^2 \,\, | \,\, d_S(\k, \k') \leq s \text{ and } d_{\mathbb{Z}^2}((x,y),(x',y')) \leq s' \}
\end{align*}
for all $(\k',x',y') \in \cX \times \mathbb{Z}^2$.
\end{dfn}

\begin{figure}[ht]
    \centering
    \includegraphics[width=.6\textwidth]{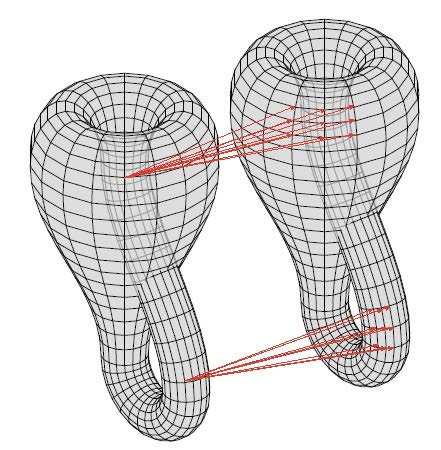}
    \caption{A visual representation of the neurons and weights in a KOL with threshold $s = 2$. Each intersection of grid lines on the two Klein bottles is a neuron. The red arrows depict the nonzero weights corresponding to the given input neuron.}
    \label{fig:KOLconnections}
\end{figure}

\begin{figure}[ht]
    \centering
    \begin{subfigure}[b]{0.2\linewidth}
        \includegraphics[width=0.98\textwidth]{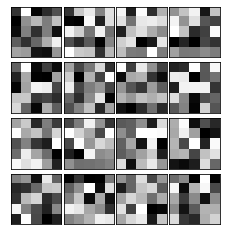}
    \end{subfigure}%
    \begin{subfigure}[b]{0.2\linewidth}
        \includegraphics[width=0.98\textwidth]{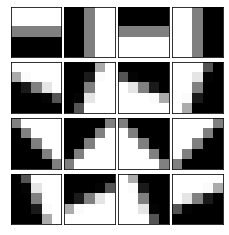}
    \end{subfigure}%
        \begin{subfigure}[b]{0.2\linewidth}
        \includegraphics[width=0.98\textwidth]{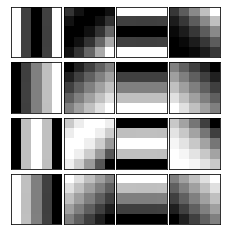}
    \end{subfigure}
    \begin{subfigure}[b]{0.2\linewidth}
        \includegraphics[width=0.98\textwidth]{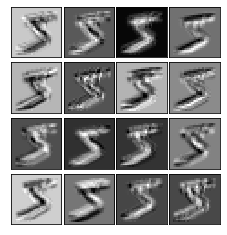}
        \caption{Normal}
    \end{subfigure}%
    \begin{subfigure}[b]{0.2\linewidth}
        \includegraphics[width=0.98\textwidth]{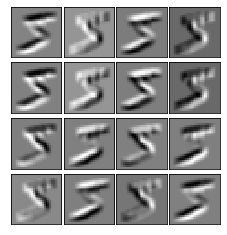}
        \caption{CF}
    \end{subfigure}%
    \begin{subfigure}[b]{0.2\linewidth}
        \includegraphics[width=0.98\textwidth]{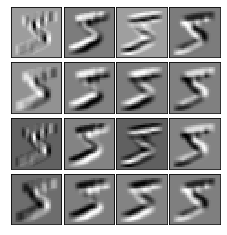}
        \caption{KF}
    \end{subfigure}
    \caption{Table of weights (top row) and activations (bottom row) for 3 convolutional layers (a),(b),(c) evaluated on a handwritten $5$ from MNIST (initial data shown in Figure~\ref{fig:guide}). The first column (a) is a normal CNN layer (NOL) trained on one epoch of MNIST data followed by 2 fully connected linear layers. This network has a testing accuracy of approximately $99\%$. Second (b) is a Circle Filters (CF) layer. Third (c) is a Klein Filters (KF) layer.}
    \label{fig:featureActivationsOn5}
\end{figure}

Next, we define a layer that localizes weights with respect to a metric on the Klein bottle $\mathcal{K}$.  See Figure~\ref{fig:KOLconnections} for a visualization of the nodes and weights. Recall that $\mathcal{K}$ is the $2$-dimensional manifold obtained from $\mathbb{R}^2$ as a quotient by the relations $(\theta_1,\theta_2) \sim (\theta_1 + 2k\pi, \theta_2 + 2l\pi)$ for $k,l \in \mathbb{Z}$ and $(\theta_1,\theta_2) \sim (\theta_1 + \pi, -\theta_2)$. The construction uses an embedding $F_\mathcal{K}$ of $\mathcal{K}$ into the vector space of quadratic functions on the square $[-1,1]^2$, motivated by the embedded Klein bottle observed in \cite{carlsson_local_2008} and its appearance in the weights of CNNs as observed in \cite{carlsson_topological_2018}.  An image patch in the embedded Klein bottle $F_\mathcal{K}(\theta_1,\theta_2)$ has a natural `orientation' given by the angle $\theta_1$. Visually, one sees lines through the center of the image at angle $\theta_1 + \pi/2$; see the top right image in Figure~\ref{fig:featureActivationsOn5}. The embedding is given by
\begin{equation} \label{eq:Kleinembedding}
F_\mathcal{K}(\theta _1, \theta_2) (x,y) = \sin(\theta _2)(\cos(\theta _1)x + \sin(\theta _1) y) + 
\cos(\theta _2) Q(\cos(\theta _1)x + \sin(\theta _1)y),
\end{equation}
where $Q(t) = 2t^2-1$. As given, $F_\mathcal{K}$ is a function on the torus, which is parameterized by the two angles $\theta _1$ and $\theta _2$.  It actually defines a function on  $\mathcal{K}$ since it satisfies $F_\mathcal{K}(\theta_1,\theta_2) = F_\mathcal{K} (\theta_1 + 2k\pi, \theta_2 + 2l\pi)$ and 
$F_\mathcal{K}(\theta _1 + \pi, - \theta _2) = F_\mathcal{K}(\theta _1, \theta _2)$.

\begin{dfn} \label{def:kleinCorr}
Let $\cX, \cX' \subset \mathcal{K}$ be two finite subsets of the Klein bottle. Let $V_i = \cX \times \mathbb{Z}^2$ and $V_{i+1} = \cX' \times \mathbb{Z}^2$ be successive layers in a FFNN. Fix a threshold $s \geq 0$.

The {\bf Klein correspondence} $C_\mathcal{K}(s) \subset \cX \times \cX'$ is defined by
$$C_\mathcal{K}(s)^{-1}(\k') = \{ \k \in \cX \,\, | \,\, d_\mathcal{K}(\k, \k')  \leq s \}$$
for all $\k' \in \cX'$, where the metric $d_\mathcal{K}$ is defined by
$$d_\mathcal{K}(\k, \k' ) = \bigg ( \int _{[-1,1]^2} \big (F_\mathcal{K}(\k)(x,y) - F_\mathcal{K}(\k')(x,y) \big )^2 dx dy \bigg )^{\frac{1}{2}}$$
for $\k, \k' \in \mathcal{K}.$

We call $V_{i+1}$ a {\bf Klein one layer (KOL)} if, for some other threshold $s' \geq 0$, the edge defining correspondence $C \subset V_i \times V_{i+1}$ is of the form
$$C = C_\mathcal{K}(s) \times C_{d,2}(s'),$$
which means that
\begin{align*}
C^{-1}(\k',x',y') &= C_\mathcal{K}(s)^{-1}(\k') \times C_{d,2}(s')^{-1}(x',y')\\
&= \{ (\k,x,y) \in \cX \times \mathbb{Z}^2 \,\, | \,\, d_\mathcal{K}(\k, \k')  \leq s \text{ and } d_{\mathbb{Z}^2}((x,y),(x',y')) \leq s' \}
\end{align*}
for all $(\k',x',y') \in \cX' \times \mathbb{Z}^2$.
\end{dfn}

We define two other layers based on the circle and the Klein bottle (Definition~\ref{def:fixedweightlayers2d}). First, we define an embedding of the circle $S^1$ into the space of functions on $[-1,1]^2$ by composing $F_{\mathcal{K}}$ with the embedding $S^1 \hookrightarrow \mathcal{K},$ $\theta \mapsto (\theta, \pi /2)$, i.e.\
$$F_{S^1}(\theta)(x,y) := F_\mathcal{K}(\theta, \pi / 2)(x,y) = \cos(\theta)x + \sin(\theta)y.$$
Now the idea is to build convolutional layers with fixed weights $\lambda$ given by discretizations of $F_{S^1}(\theta)(x,y)$ and $F_\mathcal{K}(\theta_1,\theta_2)(x,y)$. This is motivated by \cite{carlsson_topological_2018} which showed that convolutional neural networks (in particular VGG16) learn the filters $F_{S^1}(\theta)$. Instead of forcing the neural networks to learn these weights, we initialize the network with these weights. Intuitively, this should cause the network to train more quickly to high accuracy. Moreover, we choose to fix these weights during training (gradient $= 0$) to prevent overfitting, which we conjecture contributes to our observed improvement in generalization to new data. We also use discretizations of the images $F_{\mathcal{K}}(\theta_1,\theta_2)$ given by the full Klein bottle as weights, motivated by the reasoning that the trained weights observed in \cite{carlsson_topological_2018} are exactly the high-density image patches found in \cite{carlsson_local_2008} which cluster around the Klein bottle. These layers with fixed weights can be thought of as a typical pretrained convolutional layer in a network.

\begin{dfn} \label{def:fixedweightlayers2d}
Let $M = S^1$ or $\mathcal{K}$ and let $\cX \subset M$ be a finite subset. Let $V_i = \mathbb{Z}^2$ and $V_{i+1} = \cX \times \mathbb{Z}^2$ be successive layers in a FFNN. Suppose $V_{i+1}$ is a convolutional layer with threshold $s \geq 0$ (Definition~\ref{def:cnndef}). Then $V_{i+1}$ is called a {\bf Circle Features (CF) layer} or a {\bf Klein Features (KF) layer}, respectively, if the weights $\lambda_{-,(\kappa,-,-)}$ are given for $\kappa \in \cX$ by a convolution over $V_i$ of the filter of size $(2s+1) \times (2s+1)$ with values
$$\text{Filter}(\kappa)(n,m) = \int_{-1 + \frac{2m}{2s+1}}^{-1 + \frac{2(m+1)}{2s+1}}\int_{-1 + \frac{2n}{2s+1}}^{-1 + \frac{2(n+1)}{2s+1}} F_M(\kappa)(x,y) dx dy$$
for integers $0 \leq n, m \leq 2s.$
\end{dfn}

In summary, we have the following. Both $\mathcal{K}$ and $S^1$ can be discretized into a finite subset $\cX$ by specifying evenly spaced values of the angles. Given such a discretization $\cX$, the convolutional layers in a TCNN have slices indexed by $\cX$. Note that $\cX$ has a metric induced by the embedding $F_{\mathcal{K}}$ and the $L^2-$metric on functions on $[-1,1]^2$.  The {\bf COL} and ${\bf KOL}$ layers are convolutional layers with slices indexed by $\cX$ where all weights between slices whose distances in $\cX$ are greater than some fixed threshold are forced to be zero. The {\bf CF} and {\bf KF} layers are convolutional layers with slices indexed by $\cX$ and such that the weights are instantiated on the slice corresponding to $(\theta_1,\theta_2) \in \mathcal{K}$ to be the image $F_\mathcal{K}(\theta _1, \theta_2)$ discretized to the appropriate grid shape; examples of these weights are shown in Figure~\ref{fig:featureActivationsOn5}. These weights are fixed during training. See also the visual guide in Figure~\ref{fig:guide}.

\subsection{Video}\label{subsec:2dvideo}

For video, the space of features of interest is parameterized by the tangent bundle of the translational Klein bottle $\mathcal{K}^t$, denoted $T(\mathcal{K}^t)$. The translational Klein bottle $\mathcal{K}^t$ is a $3$-dimensional manifold and its tangent bundle $T(\mathcal{K}^t)$ is $6$-dimensional. These manifolds parameterize video patches \eqref{eq:videoparam} in a manner related to the Klein bottle parameterization $F_\mathcal{K}$ of 2D images patches from \eqref{eq:Kleinembedding}.

Before providing the precise definitions of $\mathcal{K}^t$ and $T(\mathcal{K}^t)$ as well as the formulas for the parameterizations, we describe the idea roughly. An image patch in the embedded Klein bottle $F_\mathcal{K}(\theta_1,\theta_2)$ has a natural `orientation' given by the angle $\theta_1$. Visually, one sees lines through the center of the image at angle $\theta_1 + \pi/2$. Given a real number $r$, there is a 2D image patch $F_{\mathcal{K}^t}(\theta_1,\theta_2,r)$ given by the translation of $F_\mathcal{K}(\theta_1,\theta_2)$ by $r$ units along the line through the origin at angle $\theta_1$, i.e.\ along the line perpendicular to the lines in the image. One can extend this image to a video that is constant in time. Videos that change in time are obtained by enlarging $\mathcal{K}^t$ to its tangent bundle $T(\mathcal{K}^t)$. The tangent bundle consists of pairs of a point $(\theta_1,\theta_2,r) \in \mathcal{K}^t$ and a vector $(u,v,w)$ tangent to $(\theta_1,\theta_2,r)$. The embedding $F_{T(\mathcal{K}^t)}$ sends such a pair to a video patch $F_{T(\mathcal{K}^t)}(\theta_1,\theta_2,r,u,v,w)$ that at time $t$ is the image $F_{\mathcal{K}^t}(\theta_1 + tu, \theta_2 + tv, r + tw).$ For example, $F_{T(\mathcal{K}^t)}(\theta_1,\theta_2,0,0,0,1)$ is the video patch that translates $F_\mathcal{K}(\theta_1,\theta_2)$ at unit speed along the line through the origin at angle $\theta_1$. Similarly, $F_{T(\mathcal{K}^t)}(\theta_1,\theta_2,0,1,0,0)$ is the video patch that rotates $F_\mathcal{K}(\theta_1,\theta_2)$ at unit speed.

Precisely, we use the following construction. Denote the coordinates on $\mathbb{R}^3 \times \mathbb{R}^3$ by the variables $(\theta_1,\theta_2,r,u,v,w)$. The variables $(\theta_1,\theta_2,r)$ parameterize $\mathcal{K}^t$ and the variables $(u,v,w)$ parameterize the tangent spaces. To be precise, $\mathcal{K}^t$ is given as the quotient of $\mathbb{R}^3$ by the relations $(\theta_1,\theta_2,r) \sim (\theta_1 + 2k\pi, \theta_2 + 2l\pi, r)$ for all $k, l \in \mathbb{Z}$ and $(\theta_1,\theta_2,r) \sim (\theta_1 + \pi, - \theta_2, -r)$, and similarly $T(\mathcal{K}^t)$ can be described as a quotient of $\mathbb{R}^3 \times \mathbb{R}^3$. We suppress further discussion of these relations because they are only significant to this work in that they are respected by the embeddings $F_{\mathcal{K}^t}$ and $F_{T(\mathcal{K}^t)}$.

Let $I = [-1,1]$. Denote by $C(I^2, I)$ the space of continuous functions $I^2 \rightarrow I$, which represent image patches at infinite resolution, and similarly denote the space of video patches by $C(I^2 \times I, I)$. The embeddings
$$F_{\mathcal{K}^t} : \mathcal{K}^t \rightarrow C(I^2, I)$$
and
$$F_{T(\mathcal{K}^t)} : T(\mathcal{K}^t) \rightarrow C(I^2 \times I, I)$$
are given by
\begin{align*}
F_{\mathcal{K}^t}(\theta_1,\theta_2,r)(x,y) &= \sin(\theta_2)(\cos(\theta_1)(x + r\cos(\theta_1)) +\sin(\theta_1)(y+r\sin(\theta_1)))\\
 &+ \cos(\theta_2)Q(\cos(\theta_1)(x + r\cos(\theta_1)) + \sin(\theta_1)(y+r\sin(\theta_1)))
\end{align*}
and
\begin{equation} \label{eq:videoparam}
F_{T(\mathcal{K}^t)}(\theta_1,\theta_2,r,u,v,w)(x,y,t) := F_{\mathcal{K}^t}(\theta_1 + tu, \theta_2 + tv, r + tw),
\end{equation}
where $Q(z) = 2z^2 - 1.$

Using the embedding $F_{T(\mathcal{K}^t)}$, we define a metric on $T(\mathcal{K}^t)$ by pulling back the $L^2$ metric on $C(I^2 \times I, I)$,
\begin{equation} \label{eq:tangentkleinmetric}
d_{T(\mathcal{K}^t)}(\k,\k') := \bigg ( \int_{I^2 \times I} \big (F_{T(\mathcal{K}^t)}(\k)(x,y,t) - F_{T(\mathcal{K}^t)}(\k')(x,y,t) \big )^2 dx dy dt\bigg )^{\frac{1}{2}}  \text{ for } \k, \k' \in T(\mathcal{K}^t).
\end{equation}

The metric $d_{T(\mathcal{K}^t)}$ allows us to define a new type of layer in a neural network. Recall the $3D$ version of the convolutional correspondence $C_{d,3}(s)$ from Definition~\ref{def:cnndef}.

\begin{dfn} {\bf (6D Moving Klein Correspondence)} \label{dfn:6dmkc}
Let $\cX, \cX' \subset T(\mathcal{K}^t)$ be two finite subsets. Let $V_i = \cX \times \mathbb{Z}^3$ and $V_{i+1} = \cX' \times \mathbb{Z}^3$ be successive layers in a FFNN. Fix a threshold $s \geq 0$.

The {\bf 6D Moving Klein correspondence} $C_{T(\mathcal{K}^t)}(s) \subset \cX \times \cX'$ is defined by
$$C_{T(\mathcal{K}^t)}(s)^{-1}(\k') = \{ \k \in \cX \,\, | \,\, d_{T(\mathcal{K}^t)}(\k, \k')  \leq s \}$$
for all $\k' \in \cX$, where the metric $d_{T(\mathcal{K}^t)}$ is defined in \eqref{eq:tangentkleinmetric}.

We call $V_{i+1}$ a {\bf 6D Moving Klein one layer (6MKOL)} if, for some other threshold $s' \geq 0$, the edge defining correspondence $C \subset V_i \times V_{i+1}$ is of the form
$$C = C_{T(\mathcal{K}^t)}(s) \times C_{d,3}(s'),$$
which means that
\begin{align*}
C^{-1}(\k',x',y',t') &= C_{T(\mathcal{K}^t)}(s)^{-1}(\k') \times C_{d,3}(s')^{-1}(x',y',t')\\
&= \{ (\k,x,y,t) \in \cX \times \mathbb{Z}^3 \,\, | \,\, d_{T(\mathcal{K}^t)}(\k, \k')  \leq s \text{ and } d_{\mathbb{Z}^3}((x,y,t),(x',y',t')) \leq s' \}
\end{align*}
for all $(\k',x',y',t') \in \cX \times \mathbb{Z}^3$. \label{dfn:sixdimcorrespondence}
\end{dfn}

There are particular submanifolds of $T(\mathcal{K}^t)$ whose corresponding video patches we conjecture to be most relevant for video classification. In 6MKOL layers, we often choose $\cX$ and $\cX'$ to be subsets of these submanifolds. One reason to do this is that discretizing the $6$-dimensional manifold $T(\mathcal{K}^t)$ results in a large number of filters, significantly bloating the size of the neural network. Indeed, discretizing $\theta_1$ and $\theta_2$ into $4$ values each, as in our Klein bottle experiments, and discretizing the other dimensions into only $3$ values produces $4^2*3^4 = 1296$ points in $\cX$. Moreover, these are videos rather than static images, so they contain many pixels: $5^3$ pixels for $5 \times 5$ video with $5$ time steps. Another reason to specialize to submanifolds is one general philosophy of this paper: Well-chosen features, rather than an abundance of features, provide better generalization due to less over-fitting.

The five $2$-dimensional submanifolds of $T(\mathcal{K}^t)$ that we choose to work with are
\begin{equation}
\begin{aligned}
\tilde{\mathcal{K}} &:= \{ (\theta_1,\theta_2,0,0,0,0) \in T(\mathcal{K}^t) \},\\ 
S_{\t}^{\pm} &:= \{ (\theta_1,\theta_2,0,0,0,\pm 1) \in T(\mathcal{K}^t) \},\\
S_{\rho}^{\pm} &:= \{ (\theta_1,\theta_2,0,\pm 1, 0, 0) \in T(\mathcal{K}^t) \}.
\end{aligned}\label{eq:fiveManifolds}
\end{equation}

Under the embedding $F_{T(\mathcal{K}^t)}$, $\tilde{\mathcal{K}}$ corresponds to the Klein bottle images held stationary in time, $S_{\t}^{\pm}$ corresponds to the Klein bottle images translating in time perpendicular to their center line, as described above, where the sign $\pm$ controls the direction of translation, and $S_{\rho}^{\pm}$ corresponds to the Klein bottle images rotating either clockwise or counterclockwise depending on the sign $\pm$.

We now define convolutional layers with fixed weights given by discretizations of the video patches corresponding to the manifolds $T(\mathcal{K}^t), \tilde{\mathcal{K}}, S_{\t}^{\pm},$ and $S_{\rho}^{\pm}$. These can be viewed as pretrained convolutional layers that would typically appear as the first layer in a pretrained video classifier. The motivation is the same as for the analogous CF and KF layers defined for images in Definition~\ref{def:fixedweightlayers2d} -- faster training and better generalization to new data due to initializing the network with meaningful filters that are fixed during training.

\begin{dfn} \label{dfn:MVideofeatureslayer}
Let $M \subset T(\mathcal{K}^t)$ be any subset, for example a submanifold such as $\tilde{\mathcal{K}}, S_{\t}^{\pm},$ or $S_{\rho}^{\pm}$ (see \eqref{eq:fiveManifolds}), or unions of such submanifolds. Let $\cX \subset M$ be a finite subset.

Let $V_i = \mathbb{Z}^3$ and $V_{i+1} = \cX \times \mathbb{Z}^3$ be successive layers in a FFNN. Suppose $V_i$ is a convolutional layer with threshold $s \geq 0$ (i.e., the edge-defining correspondence is $C_c \times C_{d,3}(s)$ as in Definition~\ref{def:cnndef}). Then $V_{i+1}$ is called a {\bf M-Features (M-F) layer} if the weights $\lambda_{-,(\kappa,-,-,-)}$ are given for $\kappa \in \cX$ by a convolution over $V_i$ of the filter of size $(2s+1) \times (2s+1) \times (2s+1)$ with values
$$\text{Filter}(\kappa)(n,m,p) =  \int_{-1 + \frac{2p}{2s+1}}^{-1 + \frac{2(p+1)}{2s+1}} \int_{-1 + \frac{2m}{2s+1}}^{-1 + \frac{2(m+1)}{2s+1}}\int_{-1 + \frac{2n}{2s+1}}^{-1 + \frac{2(n+1)}{2s+1}} F_{T(\mathcal{K}^t)}(\kappa)(x,y,t) dx dy dt$$
for integers $0 \leq n, m, p \leq 2s.$
\end{dfn}

\subsection{Gabor filters versus Klein bottle filters} \label{subsec:gabor}

The Klein bottle filters given by $F_{\mathcal{K}}(\theta_1,\theta_2)$ as in \eqref{eq:Kleinembedding} are related to Gabor filters. In fact, besides a minor difference, they are a particular type of Gabor filters. The purpose of this section is to explain this relationship. The significance of this relationship is that, while Gabor filters are commonly used in image recognition tasks, our constructions use a particular 2-parameter family of Gabor filters that is especially important, as identified by the analysis in \cite{carlsson_local_2008}. Restricting to this 2-parameter family provides a compact set of filters that can be effectively used as pretrained weights in a neural network. One may use other families of Gabor filters for this purpose, but then the question is on what basis does one choose a particular family. The Klein bottle filters are a topologically justified choice. In Section \ref{subsec:gaborresults}, we compare the performance of some other choices of Gabor filters with the Klein bottle filters.

Recall that the Gabor filters are functions on the square $[-1,1]^2$ given by
\begin{align}
g(\lambda, \omega, \psi, \sigma, \gamma)(x,y) &= e^{-\frac{x'^2 + \gamma^2 y'^2}{2\sigma^2}} \times \cos \bigg ( \frac{2\pi x'}{\lambda}+\psi \bigg ), \label{eq:gaborfilters}\\
x' &= x \cos(\omega) + y \sin(\omega), \nonumber\\
y' &= -x \sin(\omega) + y \cos(\omega). \nonumber
\end{align}
The Klein bottle filters $F_{\mathcal{K}}(\theta_1,\theta_2)$ do not taper off in intensity near the edges of the square, so we would like to remove this tapering from the Gabor filters too for a proper comparison. This is done by removing the exponential term from $g$, or equivalently, setting $\sigma = \infty$. This simultaneously removes the dependence on $\gamma$. So we have the restricted class of filters
$$g(\lambda,\omega,\psi,\infty,0)(x,y) = \cos \bigg ( \frac{2\pi}{\lambda}x'(\omega) + \psi \bigg ).$$
Given $\theta_1,\theta_2$ parameterizing the Klein bottle filter $F_{\mathcal{K}}(\theta_1,\theta_2)$, we claim that a similar Gabor filter is given by
\begin{align*}
\tilde{g}(\theta_1,\theta_2) &:= g \bigg (2 + \frac{4}{\pi}\theta_2,\,\, \theta_1,\,\, \theta_2 + \pi, \infty, 0 \bigg )\\
&= -\cos \bigg ( \frac{\pi \cdot x'(\theta_1) }{1 + \frac{2}{\pi}\theta_2} + \theta_2 \bigg ).
\end{align*}
To see the similarity between $\tilde{g}(\theta_1,\theta_2)$ and $F_{\mathcal{K}}(\theta_1,\theta_2)$, we examine the `primary circle' $\theta_2 = \pi/2$ and the `small circle' $\theta_2 = 0$. The formula for other values of $\theta_2$ interpolates along the Klein bottle between these two circles. On the primary circle, we have
\begin{align*}
\tilde{g}(\theta_1,\pi/2) &= \sin \bigg ( \frac{\pi}{2} \cdot x'(\theta_1) \bigg )\\
F_{\mathcal{K}}(\theta_1,\pi/2) &= x'(\theta_1).
\end{align*}
These are both odd functions of $x'(\theta_1)$ that are equal to $\pm 1$ at $x'(\theta_1) = \pm 1$ and are equal to $0$ at $x'(\theta_1) = 0$. Similarly, on the small circle, we have
\begin{align*}
\tilde{g}(\theta_1,0) &= -\cos \bigg ( \pi \cdot x'(\theta_1) \bigg )\\
F_{\mathcal{K}}(\theta_1,0) &= 2 \big (x'(\theta_1) \big )^2 -1.
\end{align*}
These are both even functions of $x'(\theta_1)$ that are equal to $1$ at $x'(\theta_1) = \pm 1$ and are equal to $-1$ at $x'(\theta_1) = 0$.

\section{2D Images} \label{sec:imageexperiments}

\subsection{Experiments and Results}\label{sec:results}

We conduct several experiments on the image datasets described in Section \ref{sec:data}. On individual datasets, we investigate the effect of Gaussian noise on training the TCNN Section \ref{subsec:synthetic}, the interpretability of TCNN activations Section \ref{subsec:interpretability}, and the learning rate of TCNNs in terms of testing accuracy over a number of batches in Section \ref{subsec:learningrate}. Across different datasets, we investigate the generalization accuracy of TCNNs when trained on one dataset and tested on another in Section \ref{subsec:generalizability}. We compare TCNNs to traditional CNNs in all of these domains. We also test other choices of Gabor filters versus Klein filters in the KF layers; see Section \ref{subsec:gaborresults}. All of these experiments use one or two convolutional layers, followed by a 3-layer fully connected network terminating in 10 or 2 nodes, depending on whether the network classifies digits or cats and dogs. For additional details on the neural network models, metaparameters, and train/test splits, see Section \ref{subsec:detailsofmethods-images}.

\subsubsection{Description of Data} \label{sec:data}
We perform digit classification on 3 datasets: MNIST~\cite{mnist98}, SVHN~\cite{netzer2011reading} and USPS~\cite{hull1994database}. These datasets are quite different from either other in style, while all consisting of images of digits $0$ through $9$. In particular, a human can easily identify to which dataset a particular image belongs. This makes generalization between the datasets a non-trivial task, because neural networks that train on one of the datasets will in general overfit to the particular style and idiosyncracies present in the given data, which will be inconsistent with different-looking digits from the other datasets.  The datasets come at significantly different resolutions: $28^2$, $32^2$, and $16^2$, respectively. Additionally the sizes of the datasets vary widely: roughly $7\mathrm{e}4$, $5\mathrm{e}4$, and $7\mathrm{e}3$, respectively. SVHN digits are typeset whereas MNIST and USPS are handwritten. MNIST and USPS are direct 2-D impressions with significant pre-processing already done, while the SVHN numbers are natural images of typeset digits with tilts, warping, presence of secondary digits, and other irregularities. 

We also use two collections of labeled images of cats and dogs: Cats vs. Dogs \cite{kaggleCD} (which we call Kaggle), and the cat and dog labeled images from CIFAR-10, see~\cite{krizhevsky2009learning}. Note that the Kaggle Cats vs Dogs dataset, upon our download, included several images which were empty/corrupt and could not be loaded, so the size reported here is the total number of loadable images. This seems to be typical for this dataset. These datasets contain $2.5\mathrm{e}4$ and $1.2\mathrm{e}4$ images, respectively. The resolutions of the images in each dataset are $50^2$ and $32^2$, respectively. Since we use these datasets to test the generalization accuracy of a neural network trained on one and tested on the either, we down-resolve the Kaggle data to $32^2$ to be the same size as CIFAR-10.

\begin{center}
\begin{tabular}{ c  c  c  c }
    \hline
    Dataset  &	Size  &	Dimensions  & Link\\
    \hline
    \hline
    SVHN  &	60000 &	$32^2$ & {\tiny \url{http://ufldl.stanford.edu/housenumbers}}\\
    \hline
    USPS  &	9298 &	$16^2$ & {\tiny \url{https://web.stanford.edu/~hastie/StatLearnSparsity_files/DATA/zipcode.html}}\\
    \hline
    MNIST  &	70000 &	$28^2$ & {\tiny \url{http://yann.lecun.com/exdb/mnist/}}\\
    \hline
    CIFAR-10  &	12000 &	$32^2$ & {\tiny \url{https://www.cs.toronto.edu/~kriz/cifar.html}}\\
    \hline
    Kaggle  &	24946 &	$64^2$ & {\tiny \url{https://www.kaggle.com/c/dogs-vs-cats}}\\
    \hline
\end{tabular}
\end{center}

\subsubsection{Synthetic experiments} \label{subsec:synthetic}
The central hypothesis of this work is that constraining CNNs to use the Klein bottle filters (see KF, Definition~\ref{def:fixedweightlayers2d}) and to train with respect to the Klein bottle topology on slices (see KOL, Definition~\ref{def:kleinCorr}) will provide the model with highly meaningful local features right away. So, we expect a model trained on our CF, KF, COL, and KOL layers to outperform conventional CNNs when global noise is added to the images. To test this idea, we add class-correlated Gaussian noise $N(\mu_k, \sigma_k^2)$ to our images, $\textbf{z}_k = \textbf{x}_k + \mathcal{E}_k,  \mathcal{E}_k \sim N(\mu_k, \sigma_k^2),$ where $\textbf{x}_k$ is an image of class $k$ and $\mathcal{E}_k$ is a vector of the same dimension as the image, drawn from a normal distribution. The hyperparameters $\mu_k$ and $\sigma_k^2$ hyperparameters are drawn from $N(.2,.04)$ and $\chi^2(1)\times .04$ respectively. We test our classification models when trained on $\textbf{z}$ and tested on $\textbf{x}$ and we also test the converse. Also see the end of this section for some analysis of varying the mean of $\mu_k$ and distribution of $\sigma_k^2$.

\begin{figure}[ht]
    \includegraphics[trim=10 0 60 7,clip,width=1\textwidth]{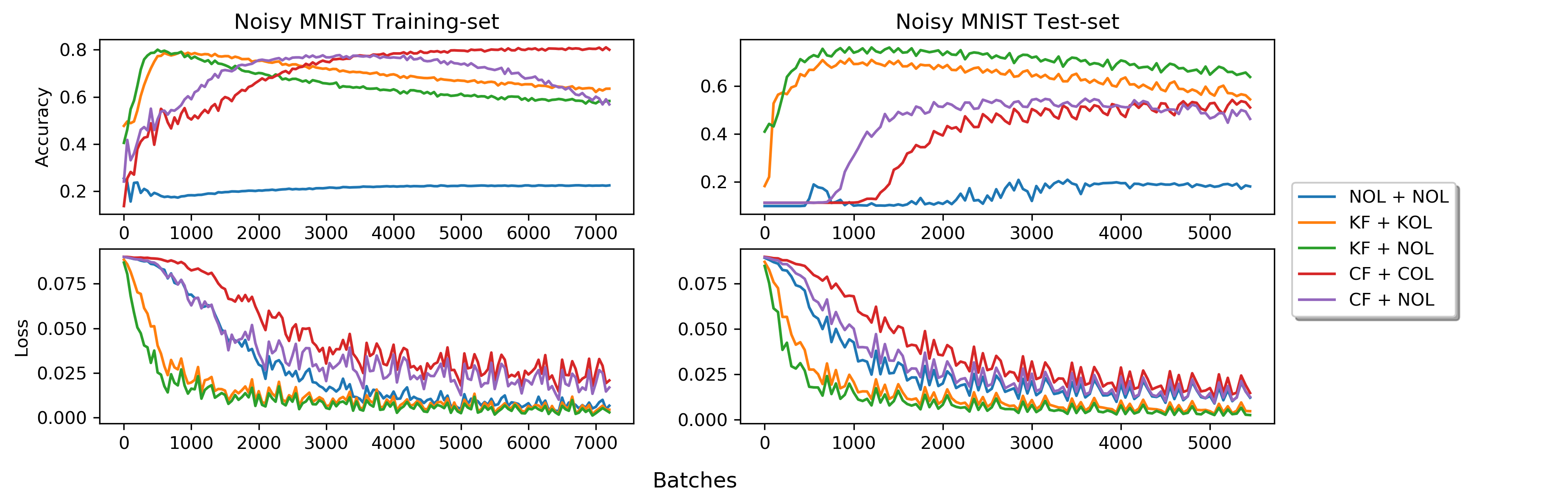}
    \caption{Two synthetic experiments on noisy MNIST data, designed to test model assumptions. The first column displays the results of the experiment where Gaussian noise is added to the training data but not the testing data. The second column displays the results of the experiment where the training data is the original MNIST data and the testing data are corrupted by Gaussian noise. The first row is testing accuracy and the second row is training loss.}\label{fig:noisyTestMnist}
\end{figure}

The results of this experiment on MNIST are displayed in Figure~\ref{fig:noisyTestMnist}. A conventional two layer network's performance is greatly deteriorated by the Gaussian noise added to the images in training or testing. In the case of noise added to the training set, we expect the CNN to learn the class-correlated $\mu_k, \sigma_k^2$, and then perform poorly on the test set without Gaussian noise. Note that Figure~\ref{fig:noisyTestMnist} includes model loss and it is clear that the 2-layer conventional CNN is classifying within the training set with high accuracy. We suspect that the TCNNs' superior performance in the noisy training set experiment is due to the fixed filters in the CF and KF layers which force a smoothing of the noise prior to the learned layers and hence generate classifiers invariant under the global noise. A similar smoothing argument explains the TCNNs superior performance in the noisy test set experiment.

All four of the TCNNs outperform conventional CNNs on training and testing. Interestingly, while our CF based systems struggle to learn at first, after many epochs of training they are best at not fitting noise. Also it is an interesting observation that the KOL and COL systems seem to learn slower but add robustness to avoiding fitting noise in lengthy training. It is important to note that this class-correlated noise is a pathological example by design, and is meant primarily to show that the topological feature and convolution layers help to prevent fitting signals outside of the edges, angles, etc. that good models are expected to fit.

\begin{figure}[hbt!]
    \begin{center}
        \includegraphics[width=1\textwidth]{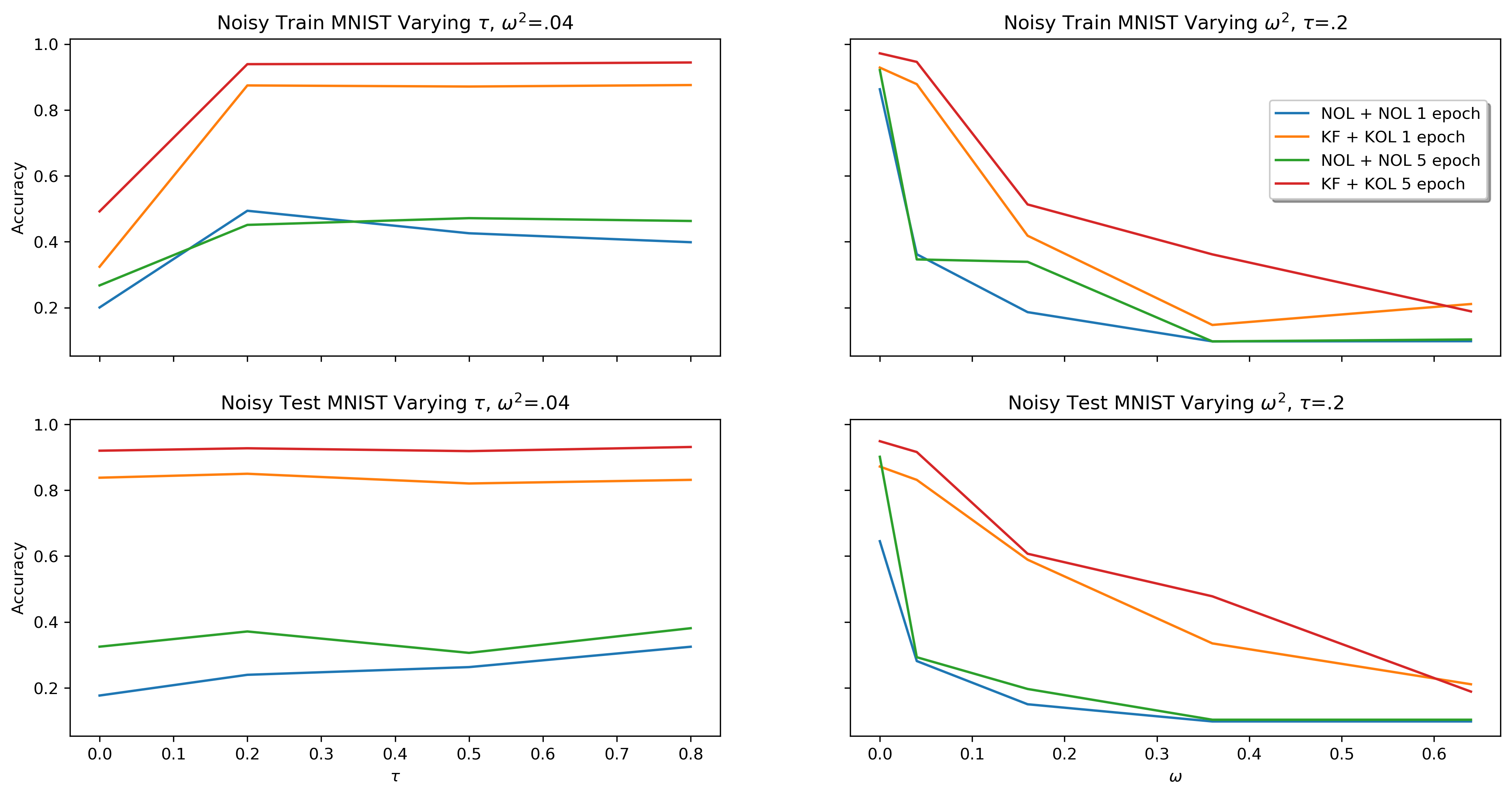}
    \end{center}
    \caption{Sweep of distributions for synthetic MNIST data varying the mean and variance from which class-noise distributions are drawn. The first column shows the tested values of $\tau$ and the second column shows the tested values of $\omega^2$. The first row shows accuracies when training on data with Gaussian noise and testing on unaltered test set. The second row shows accuracies when training on unaltered training data and testing on noise-added test data. We show results after 1 epoch and 5 epochs of training.}\label{fig:synthparamsweep}
\end{figure}

We now choose parameters  $\tau$ and $\omega$ from $0$ to $.8$ in increments of $.2$ and sample $\mu_k \sim N(\t,.04)$ and $\sigma_k^2 \sim \chi^2(1)\times \omega^2$. In the first column of Figure~\ref{fig:synthparamsweep}, we simulate distributions $\mu_k \sim N(\tau,.04)$ and $\sigma_k^2 \sim \chi^2(1)\times .04$ testing the parameter $\tau$ from $0$ to $.8$ in increments of $.2$. In the second column, we simulate distributions  $\mu_k \sim N(.2,\omega^2)$ and $\sigma_k^2 \sim \chi^2(1)\times \omega^2$ testing $\omega$ from $0$ to $.8$ in increments of $.2$.

First we examine the top row of Figure~\ref{fig:synthparamsweep} in which we train on data with Gaussian noise and test on the original data while varying $\tau$ and $\omega^2$. When $\tau$ is 0, the noise added has little class correlated signal ($\mu_k$ are small), but since $\omega^2$ is still .04, there is an uncorrelated noise component. This noise deteriorates accuracy in all models. When $\tau$ grows, the TCNN trains significantly better than the normal CNN. When $\omega^2$ is low, the models all perform similarly, and when $\omega^2$ is large, all models perform poorly. However, there is a wide range of values of $\omega^2$ for which the TCNN dramatically outperforms the CNN.

Next we examine the bottom row in which we train on the original MNIST training data and test on the Gaussian-noisy test data. Varying $\tau$ has little impact on the noisy test-set accuracy in the TCNN. As $\omega^2$ grows in the noisy test-set experiment, it increases the random component of the noise, which degrades both classifiers as expected. Again, there is a significant portion of the range of $\omega^2$ for which the TCNN is far superior to the CNN. These results are expected since we fix the convolutional weights in a sensible (topological) configuration in the KF layer. This forces the learned layers to classify on a smoothed version of the images which are more robust to the added noise.

\subsubsection{Interpretability} \label{subsec:interpretability}
We contrast the interpretability of CNNs and TCNNs by demonstrating the difference in activations between an NOL, CF, and KF (Figure~\ref{fig:featureActivationsOn5}) filtered image from MNIST. The activations shown in Figure~\ref{fig:featureActivationsOn5} correspond to the output of the first layer of CF, KF and NOL models. The NOL filters are empirically derived by training on the MNIST training data ($n=6\mathrm{e}10$). The 16 CF filters correspond to 16 evenly spaced angles on a circle and the 16 KF filters correspond to 4 evenly spaced values for each of 2 angles on the Klein bottle.

Observing the activations in Figure~\ref{fig:featureActivationsOn5}, it is empirically clear that the activations of the CF and KF layers are easier to interpret. In the case of CF, the filters are edges at various angles, and the activations show where each such orientation of an edge appears in the image. In the case of KF, filters containing interior lines are also included, and the activations reveal the presence of these interior lines in the image. The NOL learned filters and activations are, in comparison, difficult to interpret. While some of the filters seem to be finding edges, others have idiosyncratic patterns that make it difficult to assess how the combined output features contribute to a classification. The NOL model has an accuracy of approximately 99\%, so further training on MNIST is unlikely to significantly alter these filters.

\subsubsection{Rate of learning} \label{subsec:learningrate}
\begin{figure}[ht]
  \centering
    \begin{subfigure}[b]{.5\linewidth}
        \centering
        \includegraphics[width=.98\textwidth]{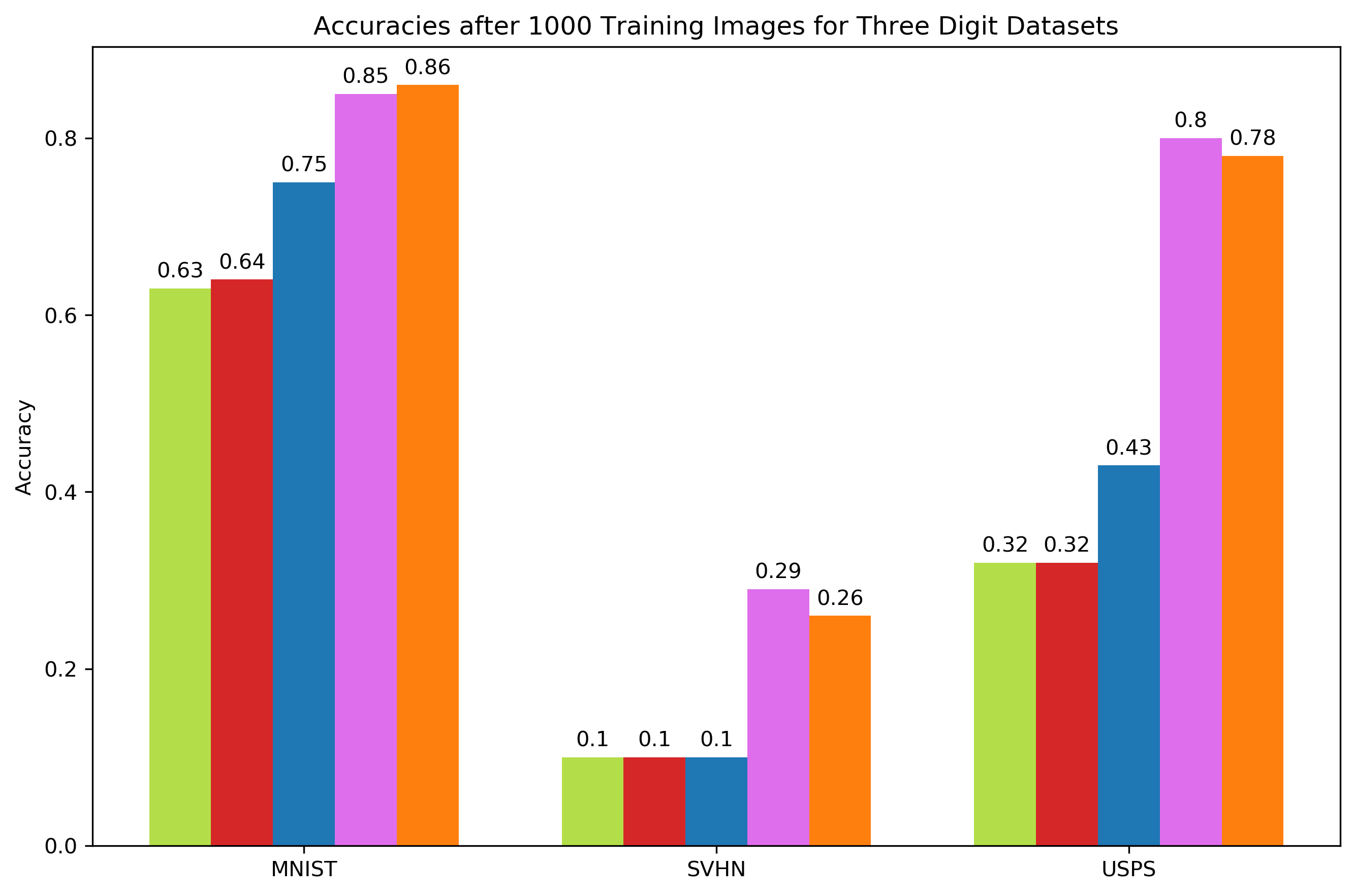}
    \end{subfigure}%
    \begin{subfigure}[b]{.44\linewidth}
        \centering
        \includegraphics[width=.98\textwidth]{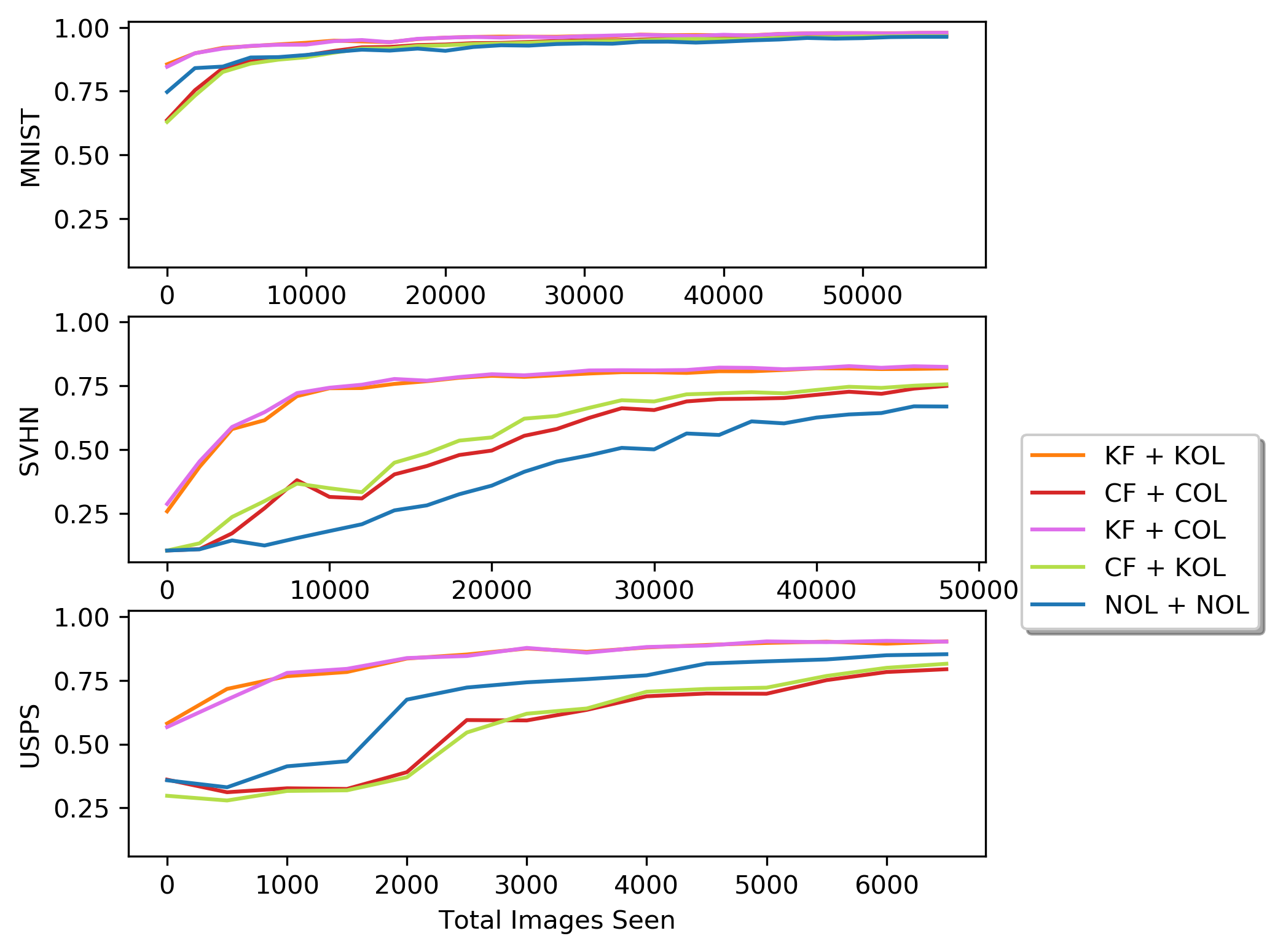}
    \end{subfigure}
    \caption{Left: Comparisons of testing accuracy after training on 1,000 images. Right: Full comparison of testing accuracy (y-axes) over a single epoch. MNIST was trained on \num[group-separator = {,}]{60000}, SVHN on \num[group-separator = {,}]{50032}, and USPS on \num[group-separator = {,}]{7291} images.}\label{fig:accRates}
\end{figure}

We find significant jumps in accuracy in the KF + KOL and KF + COL networks compared with NOL + NOL, and importantly, the networks with a KF layer achieve high accuracy earlier in training than those without, suggesting potential applications to smaller datasets for which less training can be done. See the bar chart in Figure~\ref{fig:accRates} for a comparison of testing accuracy attained from training on only 1,000 images in each dataset. The SVHN dataset, which is much richer than MNIST, and the USPS dataset, which has much lower resolution than MNIST, received larger benefits from the use of a TCNN. This suggests that the benefit of the feature engineering in TCNNs is greatest when the local spatial priors are relatively weak or hidden.

\subsubsection{Generalizability} \label{subsec:generalizability}
\begin{figure}[h]
    \begin{center}
        \includegraphics[trim=10 0 60 0,clip,width=1\textwidth]{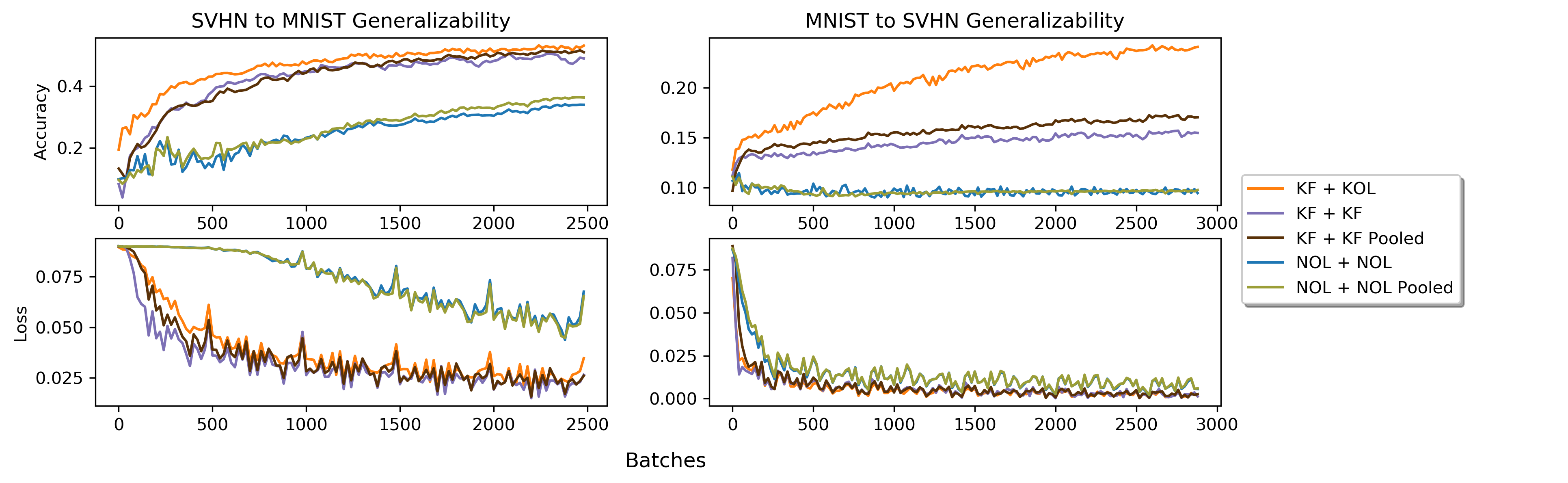}
    \end{center}
    \caption{Comparisons of testing accuracy and validation loss when generalizing from SVHN to MNIST and vice versa.}\label{fig:mnistToSvhn}
\end{figure}

\begin{figure}[h]
    \begin{center}
        \includegraphics[trim=5 0 50 0,clip,width=1\textwidth]{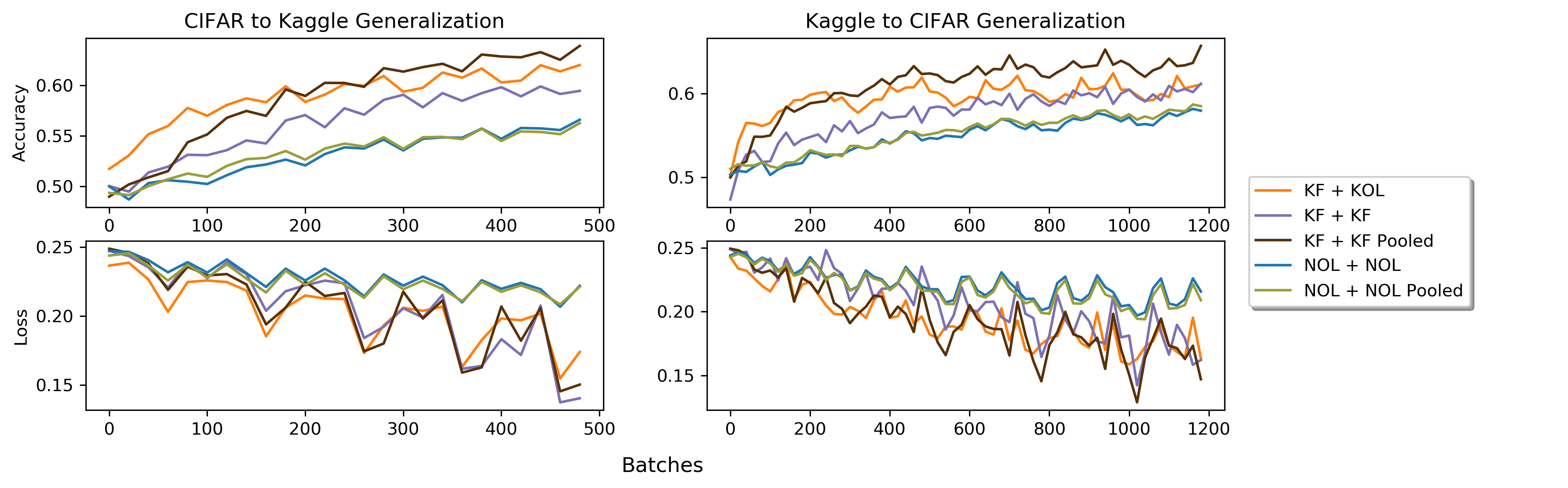}
    \end{center}
    \caption{Comparisons of testing accuracy and validation loss when generalizing from CIFAR to Kaggle and vice versa.}\label{fig:cifarToKaggle}
\end{figure}

We compare model generalizability between several models trained on either SVHN or MNIST and tested on the other, and similarly across the Kaggle and CIFAR cats vs. dogs datasets. See Figures~\ref{fig:mnistToSvhn},\ref{fig:cifarToKaggle} for a comparison of testing accuracies throughout training. The TCNNs achieve decent generalization in both the digit and cat vs dog domains. Using the KF + KOL TCNN, $30\%$ accuracy is achieved generalizing from MNIST to SVHN, and over $60\%$ accuracy is achieved generalizing from SVHN to MNIST. Contrast this with the 10\% generalization accuracy of the NOL + NOL conventional CNNs from MNIST to SVHN -- the same as random guessing. Note that the addition of a pooling layer had negligible effect on these results. Similarly, TCNNs are better than CNNs at generalizing between the Kaggle and CIFAR cats vs. dogs datasets, however the difference is less dramatic than it is for digit classification. Here, the addition of pooling layers has no impact on the generalizability of CNNs, but provides further improvement on the generalizability of TCNNs.

\subsubsection{Gabor filters versus Klein bottle filters} \label{subsec:gaborresults}

\begin{figure}[ht]
  \centering
    \begin{subfigure}[b]{.44\linewidth}
        \centering
        \includegraphics[width=.98\textwidth]{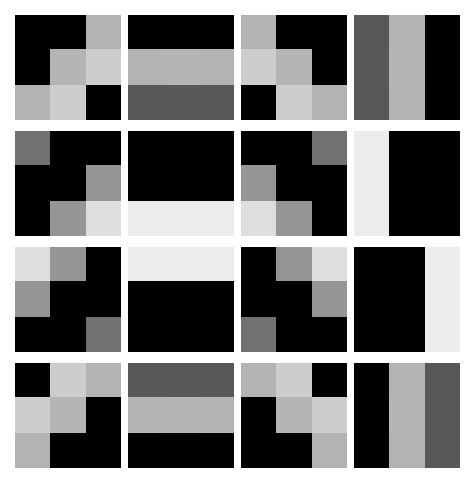}
    \end{subfigure}%
    \begin{subfigure}[b]{.44\linewidth}
        \centering
        \includegraphics[width=.98\textwidth]{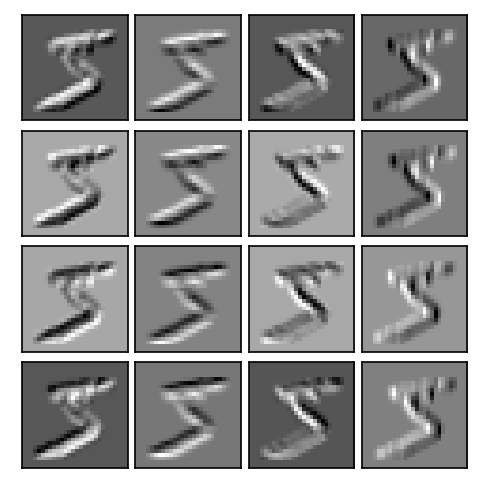}
    \end{subfigure}
    \caption{Example of a small set of gabor filters and activations on a handwritten $5$ from MNIST.}\label{fig:gaborExample}
\end{figure}

As described in Section \ref{subsec:gabor}, the Klein bottle filters given by the embedding $F_{\mathcal{K}}$ in \eqref{eq:Kleinembedding} can be viewed roughly as a subset of the $5$-parameter family of Gabor filters given by \eqref{eq:gaborfilters}. Given any subset of Gabor filters, one can make a convolutional layer analogous to the KF layer (Definition~\ref{def:fixedweightlayers2d}) where the filters are initialized to the chosen Gabor filters and frozen throughout training. In this section, we compare a KF layer with a layer instantiated with another choice of Gabor filters, given by the parameters below in \eqref{eq:gaborParams} and pictured in Figure~\ref{fig:gaborExample}. While the $2$-parameter family of Klein bottle filters is topologically justified as important for image analysis, we do not have any other means for selecting a different family of Gabor filters, and hence the $2$-parameter family in \eqref{eq:gaborParams} is necessarily somewhat arbitrary. Our heuristic motivation for this choice is that these are high contrast filters that are sparsely sampled from the two angles $\psi$ and $\omega$. The fixed parameters $\sigma$, $\lambda,$ and $\gamma$ are experimentally chosen so that the resulting $3 \times 3$ filters are high contrast. The variable parameter $\omega$ plays the same role of rotating the filter as the parameter $\theta_1$ in the parameterization $F_{\mathcal{K}}$ of the Klein bottle filters.

\begin{equation}
\sigma = 2\pi, \,\, \lambda = \pi, \,\, \gamma=\frac{\pi}{8}, \,\, \psi \in \bigg\{\frac{\pi}{4},\frac{3\pi}{4},\frac{5\pi}{4},\frac{7\pi}{4}\bigg\}, \,\, \omega\in\bigg\{\frac{\pi}{4},\frac{3\pi}{4},\frac{5\pi}{4},\frac{7\pi}{4}\bigg\}\label{eq:gaborParams}.
\end{equation}

\begin{figure}[ht]
    \begin{center}
    \begin{subfigure}{.8\textwidth}
  		\includegraphics[width=.8\linewidth]{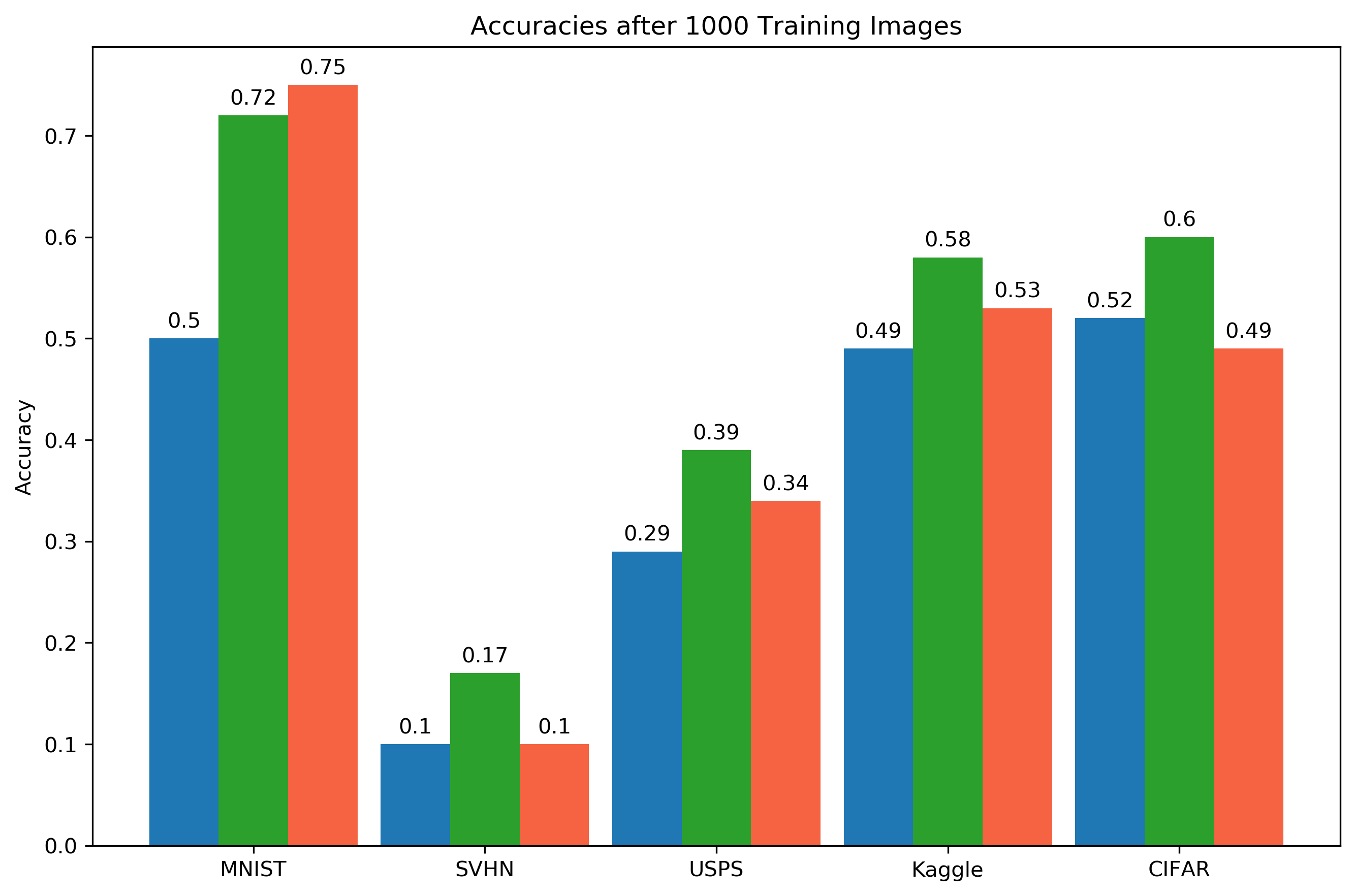}  
	\end{subfigure}
	\begin{subfigure}{.8\textwidth}
  		\includegraphics[width=.94\linewidth]{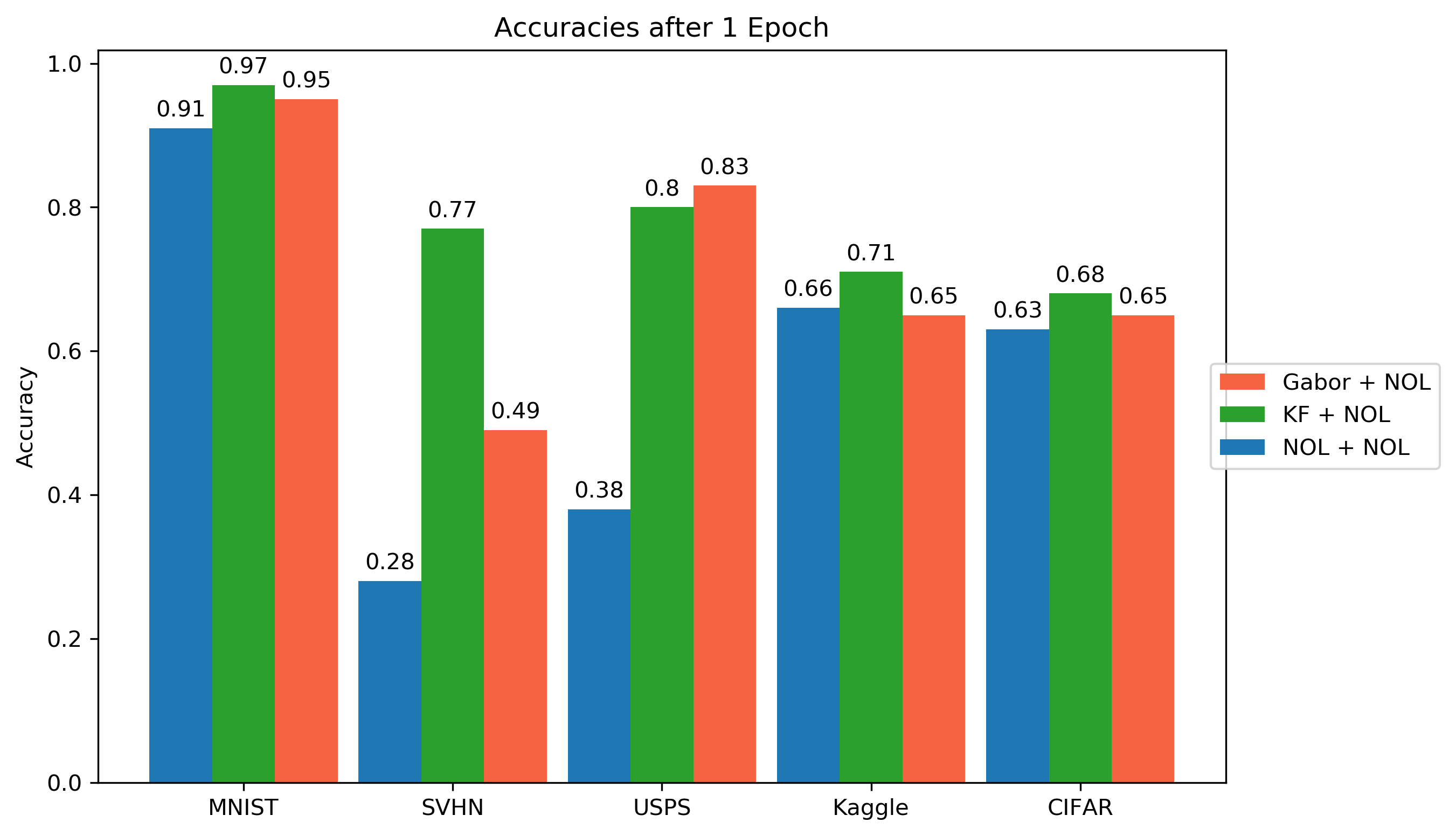}  
	\end{subfigure}
    \end{center}
\caption{Top panel: testing accuracies on 3 digit datasets and 2 cats and dogs datasets after 100 training images. Bottom panel: testing accuracies after 1 full epoch of training. The KF model consistently performs similarly or better than the Gabor filters.}\label{fig:gaborComparison1000}
\end{figure}

Using these specific Gabor filters we create a `Gabor' convolutional layer in the same fashion as the KF topological layer except with these Gabor filters as weights instead of the Klein filters. We train a Gabor + NOL convolutional network on the datasets considered in this paper and compare its accuracy to a KF + NOL network and a standard NOL + NOL network. See Figure~\ref{fig:gaborComparison1000}. We find that the Klein bottle filters perform similarly or better than the Gabor filters. This suggests that the Klein bottle filters are a good choice of Gabor filters to use in the topological layers of a TCNN, as hypothesized. Note again that some highly restrictive choice of Gabor filters must be made for this type of construction since they come in a large $5$-parameter family. Further, several of the parameters are difficult to interpret in the context of the typically small kernels of CNNs.

\subsection{Details of methods} \label{subsec:detailsofmethods-images}

\subsubsection{Train/test splits}\label{sec:traintest}
With regard to train/test splits, we perform two different types of experiment: (1) accuracy of TCNNs and CNNs are compared on a fixed dataset, and (2) either a TCNN or a CNN is trained on one dataset and validated on a second dataset to measure the capacity of the model for generalization. For type (1), we split the dataset into training and testing sets. For type (2), we train and test on the entire datasets. For type (1), the train/test splits are as follows:\\
\begin{center}
\begin{tabular}{ c  c  c }
    \hline
    Dataset  &	Train  &	Test \\
    \hline
    \hline
    MNIST  &	85\%  &	15\% \\
    \hline
    SVHN  &	80\%  &	20\% \\
    \hline
    USPS  &	80\%  &	20\% \\
    \hline
\end{tabular}    
\end{center}
For type (2), the dimension of the images is determined by the lowest resolution in the comparison, i.e., we down-resolve the higher resolution images to the lower resolution to be able to simply test generalizability.

\subsubsection{Metaparameter selection}\label{sec:metaparameters}
We choose metaparameters for each experiment that allow us to test TCNN performance against traditional CNN performance and do not seem to favor any particular model in the experiment with regard to the question at hand. Our strategy in comparing methods is to simply pick a relatively conventional, simple set of network specifications. We have selected an optimizer, batch size, learning rate, etc. on the criteria that all models are able to reasonably traverse the loss function on  which they are optimized. We select a single configuration for all models and apply this consistently throughout the experiments. While of course the precise results fluctuate depending on these choices, our findings are generally consistent no matter the configurations we chose. We expect that, since our adaptations are within the typical CNN construction, meta-parameter selection has little effect on the relative impact of choosing a TCNN over a conventional CNN.

We provide the precise metaparameter specifications and some additional detail on our reasoning below. Within each Figure in the paper, the metaparameters are the same across all models presented. The following table lists various metaparameters by Figure number.

\begin{center}
\begin{tabular}{ c  c  c  c  c  c  c }
    \hline
    Figure  &	Conv-layers  &	Conv-slices & Kernel size & LR & Batch size & Epochs\\
    \hline
    \hline
    \ref{fig:featureActivationsOn5} & $1$ & $16$ & 5 & $1\mathrm{e}{-4}$ & $100$ & $1$\\
    \hline
    \ref{fig:noisyTestMnist} and \ref{fig:synthparamsweep} & $2$ & $64$ & 3 & $1\mathrm{e}{-5}$ & $100$ & $5$\\
    \hline
    \ref{fig:accRates} & $2$ & $64$ & 3 & $1\mathrm{e}{-4}$ & $100$ & $1$\\
    \hline
    \ref{fig:mnistToSvhn} and \ref{fig:cifarToKaggle} & $2$ & $64$ & 3 & $1\mathrm{e}{-5}$ & $100$ & $5$\\
    \hline
\end{tabular}    
\end{center}

\textbf{Conv-layers.} The number of convolutional layers was chosen to be 2 for all experiments outside of Figure~\ref{fig:featureActivationsOn5}. Using 2 convolutional layers allows us to incorporate a feature layer and a correspondence layer together, e.g.\ KF+KOL. We consistently use 2 convolutional layers for uniformity throughout the paper.

\textbf{Conv-slices.} The number of slices in each convolutional layer was chosen to be 64 for all experiments outside of Figure~\ref{fig:featureActivationsOn5}. 

\textbf{Kernel size.} Figure~\ref{fig:featureActivationsOn5} contains the only experiments in which we use a kernel size of 5. This is strictly for visualization purposes, as it allows the reader to see a more nuanced picture of the circle and Klein features. All other experiments use a kernel size of 3 in each convolutional layer.

\textbf{Learning rate (LR).} An LR of $1\mathrm{e}{-5}$ or $1\mathrm{e}{-4}$ is used in each experiment. We simply choose the highest power of ten where none of the models exhibit pathological behavior.

\textbf{Batch size.} We use a batch size of 100 throughout all experiments.

\textbf{Epochs.} We do 5 epochs of training for questions of generalization and 1 epoch for questions of training speed. We choose 5 epochs for questions of generalization because training over many epochs is common in applications and often results in greater potential for over-fitting. Over-fitting should reduce generalizability, hence we trained long enough to allow for this possibility. 5 epochs allows us to compare the rate of over-fitting at different stages of training. In the cases of training speed, 1 epoch is sufficient to illustrate the comparisons since the primary interest is in the accuracy and loss over the first few batches. 1 epoch is used to create Figure~\ref{fig:featureActivationsOn5} out of convenience.

\textbf{Fully connected layers.} All experiments use 2 fully connected layers following a flattening layer. The flattening layer simply flattens the outputs of the final convolutional layer into a 1D vector. The first fully connected layer has $512$ nodes, and the second has nodes of cardinality equal to the number of classes in the output.

\subsubsection{Training} \label{sec:training}

We processed images in 100 image batches. Computing was performed on an AMD 2990WX CPU with 128GB RAM and an NVIDIA RTX 2080TI GPU.

\section{Video} \label{sec:videoexperiments}

\subsection{Experiments and Results} \label{sec:resultsVideo}

We conduct several experiments on the datasets described in Section \ref{sec:videodata}. On the UCF-101 dataset, we compare the learning rate of a TCNN in terms of testing accuracy over a number of batches in Section \ref{subsec:videolearningrate} and we compare it to the learning rate of a CNN. We also note that the TCNN achieves 70\% accuracy after 100 epochs compared to 55\% accuracy of the CNN. We also investigate the generalization accuracy of TCNNs when trained on the KTH dataset and tested on the Weizmann data in Section \ref{subsec:videogeneralizability}.

\subsubsection{Description of Data} \label{sec:videodata}

We use three datasets of videos: UCF-101~\cite{UCF101}, KTH~\cite{1334462}, and Weizmann~\cite{ActionsAsSpaceTimeShapes_pami07}. The UCF-101 dataset consists of 13320 videos in 101 classes of human actions, such as Baby Crawling, Playing Cello, Tennis Swing, etc. There are a large variety of camera motions, object scales and viewpoint, background, and lighting conditions. The videos are between 2 and 15 seconds long. The KTH dataset consists of 2391 videos of 6 types of human actions  (walking, jogging, running, boxing, hand waving, and clapping). Each video is of one of 25 different human subjects in one of a few different settings, e.g.\ outdoors and indoors, for an average length of 4 seconds. This is a significantly simpler dataset than UCF-101 due to the controlled nature of the videos e.g.\ backgrounds are homogeneous and the human actor is the only moving object. 

\begin{figure}[hbt!]
    \begin{center}
        \includegraphics[width=1\textwidth]{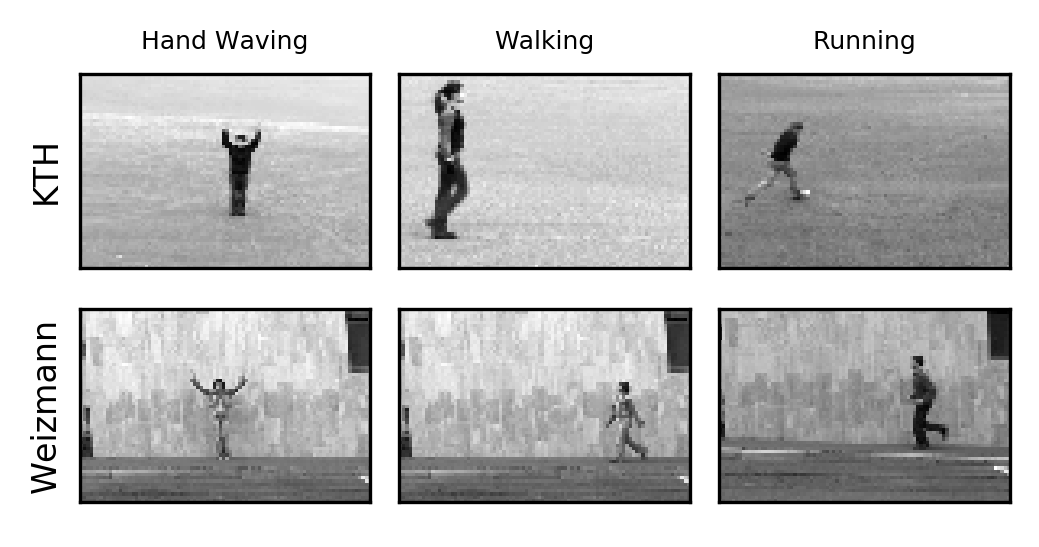}
    \end{center}
    \caption{Examples of one frame from Weizmann (top row) and KTH (bottom row) videos from the same classes (columns).}\label{fig:weizKTH}
\end{figure}

The Weizmann dataset is used to test of generalization of a network trained on the KTH dataset. Only 3 Weizmann classes have a perfect analog in KTH: handwaving, running and walking. We test generalization on these 3 classes. Example frames from Weizmann and KTH, where their classes intersect, are provided in Figure~\ref{fig:weizKTH}. There are $29$ Weizmann videos in the three classes. Both KTH and Weizmann have variable image lengths and KTH has half the frame rate of Weizmann. We choose to take about 1 second of each video at 25fps on which to perform classification.

\begin{center}
\begin{tabular}{ c  c  c  c }
    \hline
    Dataset  &	Size  &	Dimensions  & Link\\
    \hline
    \hline
    UCF-101  &	13320 & 320x240x25fps	 & https://www.crcv.ucf.edu/data/UCF101.php \\
    \hline
    KTH  &	 2391 & 160x120x25fps & https://www.csc.kth.se/cvap/actions/ \\
    \hline
    Weizmann & 29	 & 180x144x50fps	 & http://www.wisdom.weizmann.ac.il/$\sim$vision/SpaceTimeActions.html\\
    \hline
\end{tabular}
\end{center}

\subsubsection{Accuracy and rate of learning on a ResNet} \label{subsec:videolearningrate}

We train 12 layer ResNet classifiers as in \cite{he2016deep} on the UCF-101 dataset. One of these classifiers is a conventional CNN whose convolutional layers operate on all 3 dimensions of the video. The other classifier is a TCNN that is identical to the CNN except for the first layer in which we use a M-F layer (Definition~\ref{dfn:MVideofeatureslayer}) instead of a NOL. The M-F layer is of type $M = \tilde{\mathcal{K}} \cup S_{\t}^{\pm} \cup S_{\rho}^{\pm}$ (see \eqref{eq:fiveManifolds}),
that is, we use the features given by translating the Klein bottle in time, rotating the Klein bottle in time, and holding the Klein bottle still in time -- see the discussion above Definition~\ref{dfn:MVideofeatureslayer}.

\begin{figure}[hbt!]
    \begin{center}
        \includegraphics[width=.6\textwidth]{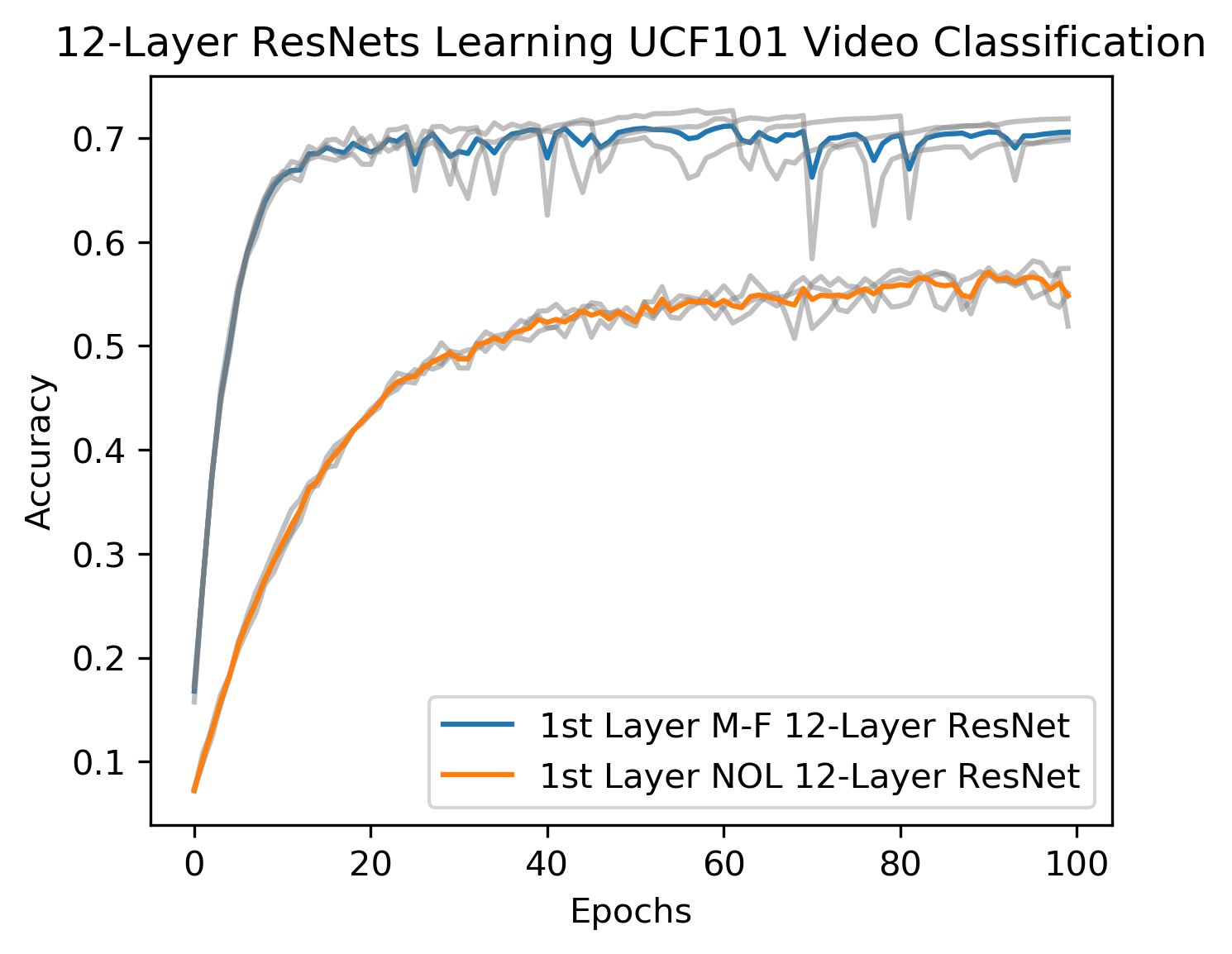}
    \end{center}
    \caption{Testing accuracy of a 12 layer ResNet trained on UCF-101.}\label{fig:ucfHighAccClassAllRuns}
\end{figure}

\begin{figure}[hbt!]
    \begin{center}
        \includegraphics[width=1\textwidth]{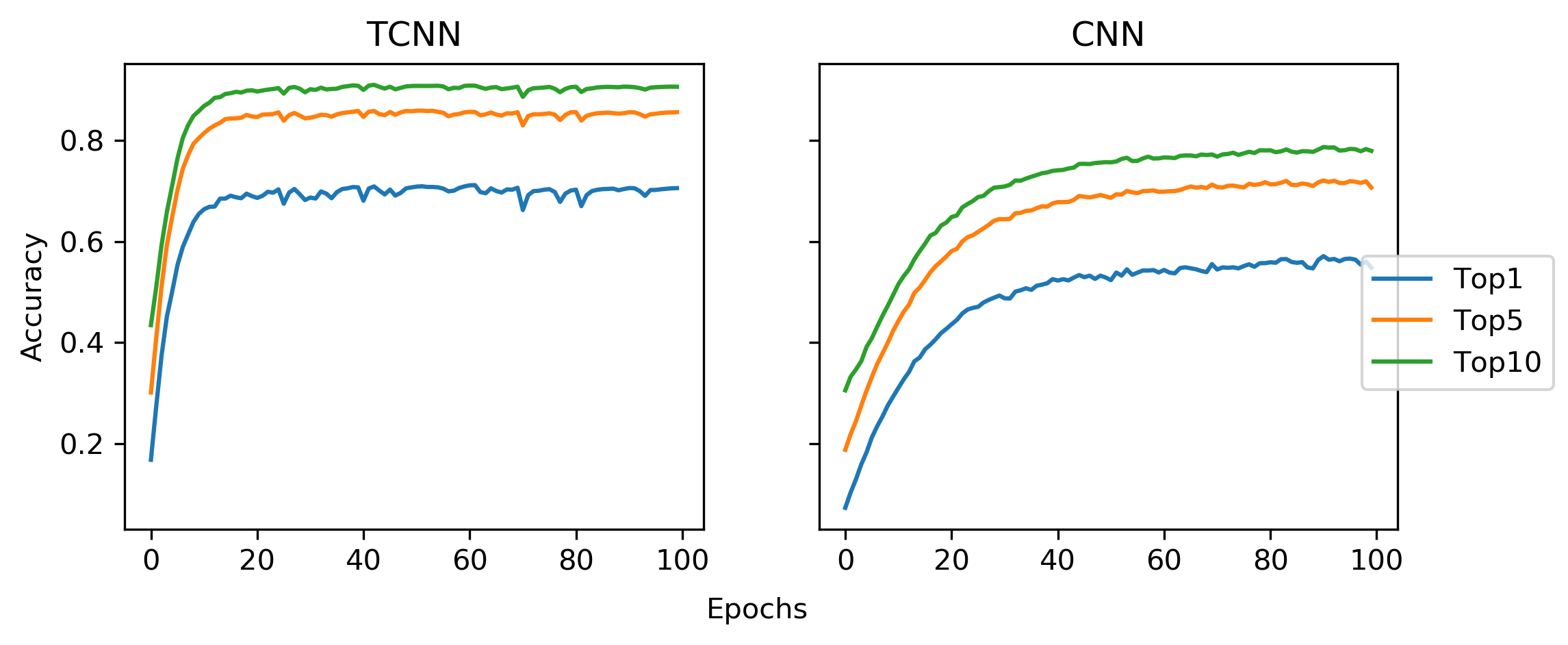}
    \end{center}
    \caption{Top 1, 5, and 10 testing overall classification accuracies as a function of epochs for both the TCNN ResNet (left) and the CNN ResNet (right).}\label{fig:ucfTopAccComparison}
\end{figure}

Our main result is that, when using the M-F layer as the first layer instead of NOL, we find a significant increase in final accuracy as well as a much faster learning rate in the initial epochs; see Figure~\ref{fig:ucfHighAccClassAllRuns}. In Figure~\ref{fig:ucfTopAccComparison}, we plot the best class accuracy, best five, and best ten, as a function of training epochs.

We also perform the same experiment on the KTH dataset. The results are plotted in blue and green in Figure~\ref{fig:kthVsWeizmannGen}. Again, the TCNN achieves high accuracy more quickly and remains higher throughout all 200 epochs of training. The difference in final accuracy between the TCNN and the CNN is not as dramatic as we observe on the UCF-101 dataset, because the KTH dataset experiences less variation between videos, so it is easier for the CNN to learn meaningful features without the help of the M-F layer.

\subsubsection{Generalizability} \label{subsec:videogeneralizability}

We compare generalizability of a TCNN and a CNN model trained on the KTH dataset and tested on the Weizmann dataset. Both of these models are ResNets, and the TCNN is the same as the CNN except with the first layer a M-F layer instead of a NOL layer, as described in Section \ref{subsec:videolearningrate}. The TCNN generalizes from KTH to Weizmann with a testing accuracy of about 65\%, whereas the CNN achieves only about 52\% generalization accuracy.

\begin{figure}[hbt!]
    \begin{center}
        \includegraphics[width=.8\textwidth]{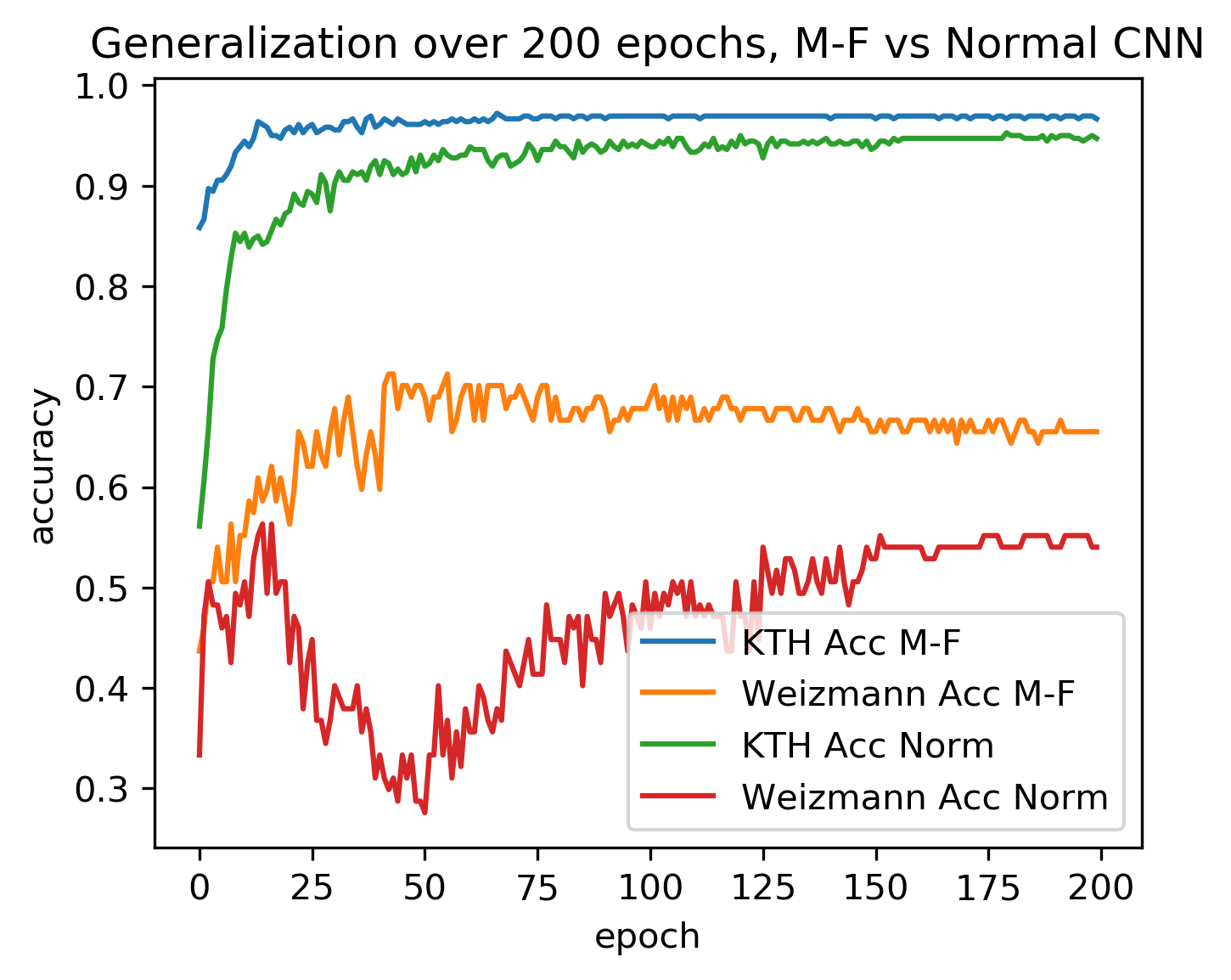}
    \end{center}
\caption{Testing accuracy of a TCNN and a CNN trained on the KTH dataset and tested on the Weizmann dataset.}\label{fig:kthVsWeizmannGen}
\end{figure}

The full results are shown in Figure~\ref{fig:kthVsWeizmannGen}. Observe that the generalization accuracy of the CNN takes a large dive at around 50 epochs, which suggests that it is initially learning features specific to the KTH dataset that do not generalize well, and then as it continues to train it learns more meaningful features that do generalize. We conjecture that the dip in accuracy on the Weizmann dataset occurs during the transition from learning KTH-specific artifacts to learning meaningful features. On the other hand, the TCNN immediately begins learning meaningful features that generalize well. This is the desired effect of the M-F layer -- it biases the learner to use significant features in the video right away, instead of initially learning artifacts particular to the dataset.

\subsection{Details of methods}

\subsubsection{Train/test splits}
In our video experiments, we only have a train/test split for UCF-101. KTH and Weizmann videos were only used in the test of generalization, so for these datasets we use the entire set of KTH in training and the entire set of Weizmann to test (in the three common classes). The UCF-101 split is as follows:
\begin{center}
\begin{tabular}{ c  c  c }
    \hline
    Dataset  &	Train  &	Test \\
    \hline
    \hline
    UCF-101  &	67\%  &	33\% \\
    \hline
\end{tabular}    
\end{center}

\subsubsection{Metaparameter selection}

\begin{center}
\begin{tabular}{ c  c  c  c  c  c  c }
    \hline
    Figure  &	Conv-layers  &	Conv-slices & Kernel size & LR & Batch size & Epochs\\
    \hline
    \hline
    \ref{fig:kthVsWeizmannGen} & $3$ & $20,80,160$ & $(5\times 11\times 11),(5\times 11\times 11),(3\times 9\times 4)$ & $3\mathrm{e}{-3}$ & $100$ & $200$\\
    \hline
    \ref{fig:ucfHighAccClassAllRuns} \& \ref{fig:ucfTopAccComparison} & $12$ & $180\times 9,360\times 3$ & $(5, 11, 11),(1, 5, 5),(1, 3, 3),(3, 3, 3)$ & $1\mathrm{e}{-5}$ & $100$ & $5$\\
    \hline
\end{tabular}    
\end{center}

Our video experiments differ from image experiments in that we use kernels which are not square (cubic). To indicated the size of a 3-dimensional kernel, we use the notation $(t,y,x)$, where $t$ is the size in the time dimension, and $x,y$ are spatial sizes in their respective dimensions.

Notably, the ResNet12 residual network which produced our highest accuracy classifiers for UCF-101 (Figures~\ref{fig:ucfHighAccClassAllRuns} and \ref{fig:ucfTopAccComparison}), has a significantly different form from all other models in this work, both in depth and in the use of a residual block. We choose a simple block architecture:
\begin{table}
\begin{center}
\begin{tabular}{ c }
	Block($\nu, x$):\\
    \hline
    \hline
    $y=\mathrm{Convolution}(\nu)(x)$\\
    \hline
    $y=\mathrm{Convolution}(\nu)(y)$\\
    \hline
    $y=\mathrm{Batch}(y)$\\
    \hline
    $y=\mathrm{Upsample}(y)$\\
    \hline
	$x+y$  \\ 
    \hline
\end{tabular}
\caption{Residual block for our 12-convolutional layer ResNet where $x$ is an input tensor, data passes from top to bottom, so $x+y$ is the returned tensor.}  
\end{center}
\end{table}  
where each block has two convolutional layers with a kernel size $\nu$. In order to add the input to the output, we need tensors of matching dimension, so we use Upsample($\cdot$) which is a trilinear upsampler and Batch($\cdot$) which is performs batch normalization. We indicate a 3-dimensional pooling layer by Pool($\cdot$). Our residual network then has the following structure:

\begin{table}
\begin{center}
\begin{tabular}{ c c }
	$\nu$ & ResNet12($x$):\\
	\hline
	\hline
	~ & $y={M-F}_1(x)$\\
	\hline
	(1, 5, 5) & y=Block$(\nu,y)$\\
	\hline
	(1, 3, 3) & y=Block$(\nu,y)$\\
	\hline
	(3, 3, 3) & y=$\mathrm{Pool} \left(C_d(\nu)(y)\right)$\\
	\hline
	~ & y=Batch($y$)\\
	\hline
	(3, 3, 3) & y=$C_d(\nu)(y)$\\
	\hline
	~ & y=Batch($y$)\\
	\hline
	(3, 3, 3) & y=$\mathrm{Pool} \left(C_d(\nu)(y)\right)$\\
	\hline
	~ & y=Batch($y$)\\
	\hline
	~ & y=FC($y$)\\
	\hline
	~ & y=Batch($y$)\\
	\hline
	~ & $y=\mathrm{softmax}(y)$\\
	\hline
\end{tabular}
 \caption{Structure of our 12-convolutional layer ResNet with order of opperations indicated by descending through the table. The first column of the table indicates the kernel size of a convolutional layer or layers in a residual block.}  
\end{center}
\end{table}

\subsubsection{Training}

We processed videos in 50 (UCF-101) or 100 (KTH vs Weizmann) image batches. Computing was performed on an AMD 2990WX CPU with 128GB RAM and an NVIDIA RTX 2080TI GPU.

\section*{Acknowledgments}
Research has been partially supported by the Army Research Office (ARO) Grant \# W911NF-17-1-0313 (VM), and the National Science Foundation (NSF) Grants \# MCB-1715794 (EL, VM), DMS-1821241 (VM), DMS-1903023 (BF). GC was supported by Altor Equity Partners AB through Unbox AI (www.unboxai.org). 
We would like to acknowledge Rickard Brüel Gabrielsson and Michael McCabe for helpful conversations.

\bibliography{topnet}

\end{document}